\documentclass[11pt]{article}
\usepackage{amsmath,amssymb,amsfonts}
\usepackage{algorithmic}
\usepackage{graphicx}
\usepackage{textcomp}
\usepackage[dvipsnames]{xcolor}
\usepackage[margin=1in]{geometry}
\usepackage{url}
\usepackage{wrapfig}
\usepackage{marvosym}
\usepackage{enumerate}
\usepackage[font=small,labelfont=bf]{caption} 
\usepackage{tikz}
\usepackage{tkz-fct}
\usepackage{hyperref} 
\usepackage{fancyhdr} 
\usepackage{lastpage} 
\usepackage{extramarks} 
\usepackage{listings} 
\usepackage{courier} 
\usepackage{lipsum} 
\usepackage{booktabs}
\usepackage{bm}
\usepackage{setspace}
\usepackage{authoraftertitle}
\usepackage{parallel}
\usepackage{enumerate}
\usepackage{multicol}
\usepackage{tabu}                  
\usepackage{colortbl}
\usepackage{rotating}
\usepackage{multirow}
\usepackage{makecell}
\usepackage{bbm}
\usepackage[utf8]{inputenc}
\usepackage[english]{babel}
\usepackage{mathtools}
\usepackage{subfig}
\usepackage{cite}
\usepackage{siunitx}
\usepackage{mathbbol}

\newtheorem{theorem}{Theorem}
\newtheorem{definition}{Definition}
\newtheorem{corollary}{Corollary}[theorem]
\newtheorem{lemma}{Lemma}

\newtheorem{remark}{Remark}

\def\BibTeX{{\rm B\kern-.05em{\sc i\kern-.025em b}\kern-.08em
    T\kern-.1667em\lower.7ex\hbox{E}\kern-.125emX}}

\newenvironment{spmatrix}[1]
{\def\mysubscript{#1}\mathop\bgroup\begin{pmatrix}}
	{\end{pmatrix}\egroup_{\textstyle\mathstrut\mysubscript}}

\newcommand{\eq}[2]{\begin{equation}\label{#1} #2 \end{equation}}
\newcommand{\al}[1]{\begin{align*} #1 \end{align*}}

\newcommand{\rf}[1]{\eqref{#1}}
\newcommand{\eqs}[2]{\begin{equation} \label{#1}\begin{split} #2\end{split} \end{equation}}

\newcommand{\blue}{\color{blue}}

\newcommand{\vb}[1]{\langle #1\rangle}
\newcommand{\norm}[1]{\left\lVert#1\right\rVert}

\renewcommand{\a}{{\alpha}}
\renewcommand{\v}{\varsigma }

\newcommand{\w}{\mathbf{w}}
\newcommand{\e}{\epsilon}
\newcommand{\ve}{\varepsilon}
\newcommand{\tr}{\nabla}
\newcommand{\td}{\tilde}
\newcommand{\ee}{\mathbf{e}}
\newcommand{\g}{\gamma }

\newcommand{\C}{C_1}
\newcommand{\s}{\sigma }
\renewcommand{\b}{\beta}
\renewcommand{\k}{\kappa}
\renewcommand{\o}{\mathbf{\omega }}
\newcommand{\D}{\Delta}

\newcommand{\trs}{^\intercal}

\renewcommand{\l}{{\delta}}
\newcommand{\x}{\mathbf{x} }
\newcommand{\y}{\mathbf{y} }
\newcommand{\X}{\mathbf{X} }

\newcommand{\E}{\mathbb{E}}
\newcommand{\var}{\mathrm{Var}}

\newcommand{\sumbk}{ \sum_{k=1}^N\a_k}
\newcommand{\sumbks}[1]{ \sum_{k=1}^N\a_k{#1}}
\newcommand{\sumbs}[1]{ \sum_{i=1}^N\a_i{#1}}
\newcommand{\sumbss}[1]{ {#1}\sum_{i=1}^N \a_i}
\newcommand{\sumbkss}[1]{ {#1}\sum_{k=1}^N\a_k}

\newcommand{\itt}{_{i,t}}
\renewcommand{\ij}{_{i,j}}
\newcommand{\jt}{_{j,t}}
\newcommand{\kt}{_{k,t}}

\newcommand{\ub}{ K_2d\td{B}_t} 
\newcommand{\js}{C_2} 
\newcommand{\tg }{\g} 
\newcommand{\hb }{\hat{\beta}} 
 
\newcommand{\hz }{\hat{\w}_t} 
 
\newcommand{\hl}{\hat{\lambda}_2}
\newcommand{\R}{\mathbb{R}}
\renewcommand{\L}{\Lambda}

\renewcommand{\O}{\Omega}
\newcommand{\oi}{\O^{-1}}
\newcommand{\tw}{\w^*}
\newcommand{\trc}[1]{\text{tr}{(#1)}}

\def\*#1{\bm{#1}}
\def\@#1{\mathcal{#1}}

\def\QEDmark{\ensuremath{\square}}
\def\proof{{\em Proof: }}
\def\endproof{\hfill\QEDmark}

\providecommand{\keywords}[1]{\textbf{\textit{Keywords:}} #1}

\makeatletter
\def\endthebibliography{%
	\def\@noitemerr{\@latex@warning{Empty `thebibliography' environment}}%
	\endlist
}
\makeatother

\begin{document}
	\title{Distributed Networked Learning with Correlated Data}	
	\author{Lingzhou Hong, Alfredo Garcia, and Ceyhun Eksin
		\thanks{This work was supported by NSF ECCS-1933878, NSF CCF-2008855, and Grant AFOSR-15RT0767. }
		\thanks{The authors are with Department of Industrial \& Systems Engineering, Texas A\&M University, College Station, TX 77843. E-mail: \{hlz, alfredo.garcia, eksinc\}@tamu.edu}
}


\date{}

	\maketitle

\begin{abstract}
	We consider a distributed estimation method in a setting with heterogeneous streams of correlated data distributed across nodes in a network. In the considered approach, linear models are estimated locally (i.e., with only local data) subject to a network regularization term that penalizes a local model that differs from neighboring models. 
	We analyze computation dynamics (associated with stochastic gradient updates) and information exchange (associated with exchanging current models with neighboring nodes). We provide a finite-time characterization of convergence of the weighted ensemble average estimate and compare this result to {\em federated} learning, an alternative approach to estimation wherein a single model is updated by locally generated gradient updates. This comparison
	highlights the trade-off between speed {\em vs} precision: while model updates take place at a faster rate in federated learning, the proposed networked approach to estimation enables the identification of models with higher precision. We illustrate the method's general applicability in two examples: estimation of a Markov random field using wireless sensor networks and modeling prey escape behavior of birds based on a publicly available dataset. 
\end{abstract}



%
%
%
%
%
%
%
%
%


\section{Introduction}
\label{sec:introduction}
The ever-growing size and complexity of data create scalability challenges for storage and processing. In certain application domains, data cannot be stored or processed in a single location due to geographical constraints or limited bandwidth. In such cases, a distributed architecture for data storage or processing relying on a network of interconnected computers (not necessarily in the same physical location) is often required. 

In this paper, we consider the problem of estimating a linear model in real-time based upon heterogeneous streams of correlated data that are distributed across nodes in a network. Since data streams are neither independent nor identically distributed, a model estimated based exclusively on local data may be of arbitrarily low precision. When data centralization is not feasible nor desirable (e.g., due to privacy concerns), the challenge for a distributed approach to estimation consists of identifying algorithmic solutions with low overhead that guarantee improved precision for models obtained exclusively with local data.
In this paper, we consider an approach to a distributed estimation that successfully addresses these concerns.

In the proposed approach, locally estimated linear models are updated in response to new gradient estimates for either a local loss measure (generalized least squares) or a network regularization function. Such function penalizes a local model in a manner proportional to the distance to other neighboring local models. The regularization-based updates require periodic exchanges of locally identified models amongst neighbors, a task with relatively low communication overhead when models are not high dimensional.

In the first part of the paper, we analyze computation dynamics (associated with stochastic gradient updates) and information exchange (associated with exchanging current models with neighboring nodes to compute the gradient of network regularization).
To undertake the analysis, we use a continuous-time approximation of the underlying stochastic difference equations, which allows the use of Ito's calculus.

In Theorem 1, we provide a finite-time characterization of convergence of the weighted ensemble average estimate using an upper bound on a regularity (or dispersion)  measure of the local models.
Such upper bound is influenced by the local model exchanging rate and the network connectivity degree.
The regularity measure can be made arbitrarily small for a large enough value of a network regularization parameter. In this case, the weighted ensemble average model is also arbitrarily close to any locally estimated model.

In Theorem 2, we provide a finite-time characterization of convergence of the weighted ensemble average model error. We show the rate of convergence is determined by {\em smallest} strong convexity parameter across all nodes (i.e. $\kappa>0$) and the {\em slowest} data rate (i.e. $\mu>0$).
The asymptotic error is increasing in the {\em worst case} condition number $\frac{\eta}{\kappa}>1$ where $\eta>0$ is the maximum value of the Lipschitz (gradient smoothness) constants associated with each node. 
The asymptotic error is also increasing in data rate imbalance, i.e., $\frac{\mu'}{\mu}>1$ where $\mu'>0$ is the {\em fastest} data rate across all nodes.
This characterization has no dependence on the dimension of the models. Hence, the characterization remains valid for higher-dimensional models as long as the worst-case condition number is bounded (i.e., $\kappa>0$ is bounded away from zero and $\eta< \infty$).

We compare this performance characterization with that of an alternative approach to estimation known as {\em federated} learning (FL) (see, e.g., \cite{FL6, FL3}). A single model stored in a shared (centralized) parameter server is updated with locally generated gradient updates in this approach. While model updates take place at a faster rate in FL, the proposed networked approach to estimation enables the identification of models with higher precision. This is formalized in two corollaries to Theorem 3. 

In the first corollary, a large enough network, i.e., one with at least $N>\sqrt{\frac{\mu ^{\prime }\eta }{\mu \kappa }}$ nodes in a connected topology, is shown to asymptotically exhibit higher average model precision. A networked estimation approach is also more robust to heterogeneity in noise distribution. With increasing disparities in noise variance, the FL approach is more vulnerable to noise. For example, if nodes with faster data rates are also noisier, the identified model estimate will inevitably be noisy. In the second corollary we show that the networked approach is guaranteed to outperform FL estimates when a measure of heterogeneity in noise variance across nodes (i.e. $\frac{\bar{\sigma}^{2}}{\min \sigma _{k}^{2}}$) exceeds the threshold $\sqrt{\frac{\mu ^{\prime }\eta }{\mu \kappa }}$. These corollaries highlight a trade-off between speed {\em vs} precision: while model updates take place at a faster rate in federated learning, the proposed networked approach to estimation enables the identification of models with higher precision\footnote{We note that the preliminary version of this study appeared as a conference publication \cite{hong2020distributed}. The model described here incorporates heterogeneity in the speed of data processing across nodes and does not commit to a particular choice of weights for updates based on models received from neighbors, unlike the preliminary model\cite{hong2020distributed}. We present convergence results (Theorems 1 and 2) similar to \cite{hong2020distributed} with these generalizations. These generalizations provide additional insights into the implications of data rate imbalance combined with the heterogeneity of data discussed above. In addition to the model generalization, we provide an analytical comparison of the method's performance with FL (Theorem 3, and Corollaries 1 and 2).}.

This paper is related to several strands of the literature. In a ``divide and conquer" approach to distributed data (see, e.g., \cite{Zhang2015}, \cite{Predd} and \cite{Zhang2013}), individual nodes implement a particular learning algorithm to fit a model for their assigned data set and {\em upon each machine identifying a model}, an ensemble (or global) model is obtained by averaging individual models. This is similar to {\em ensemble} learning (see, e.g., \cite{Mendes}), which refers to methods that combine different models into a single predictive model. For example, bootstrap aggregation (also referred to as ``bagging") is a popular technique for combining regression models from {\em homogeneously} distributed data.\footnote{ A careful selection of weights for computing the average model ensures a reduction of estimation variance along with other desirable properties, see, e.g., \cite{Hansen}, \cite{Liu2016}).} 
While a ``divide and conquer" approaches coupled with a model averaging step can significantly reduce computing time and lower single-machine memory requirements, it relies on a single synchronized step (i.e., computing the ensemble average), which is executed {\em after all} machines have identified a model. In contrast, the approach considered in this paper deals with {\em asynchronous} real-time estimation and regularization for {\em heterogeneous} and {\em correlated} data streams.

The considered scheme is related to the literature on consensus optimization (see, e.g., \cite{nedic2009distributed,Lian,Yin})
and the recent work on finding the best common linear model in convex machine learning problems \cite{he2018cola}. However, as we shall show, the proposed approach can not be interpreted as being based upon averaging local models as in consensus-based optimization. The algorithms proposed in \cite{Lian} and \cite{Yin} are designed for {\em batch} data while our approach deals with {\em streaming} data. For example, in \cite{Lian}, gradient estimation noise is assumed independent and homogeneous, while in our approach, gradient estimation noise is {\em correlated} and {\em heterogeneous}. In addition, the algorithms proposed in \cite{Lian} and \cite{Yin}, every node is {\em equally likely} to be selected at each iteration to update its local model. 
In contrast, in our approach, data streams are heterogeneous so that certain nodes have faster data streams and thus are more likely to update their models at any point in time.
A network regularization penalty for networked learning has been analyzed in a series of papers by \cite{Sayed_1, Sayed_2, Sayed_4, Sayed_5, 2015distributed, garcia2020,kar2013distributed}. In contrast to these papers, we consider a setting with heterogeneous nodes with {\em correlated} data streams {\em asynchronously} updating their respective models at {\em different rates} over time.

Finally, the paper is related to the literature of distributed algorithms to solve linear algebraic equations (such as those associated with generalized least squares) over multi-agent networks (see, e.g., \cite{Mou, Liu_2015, Wang}). However, unlike these papers, we examine the consequences of heterogeneous and correlated noise in distributed generalized least squares estimation in the present paper.

The contributions of this paper are as follows. We develop a distributed estimation scheme that accounts for heterogeneous and correlated distributed datasets and heterogeneity in data processing speed by the nodes. We provide a finite-time characterization of convergence of the weighted ensemble average that captures the performance gap between the centralized and weighted average ensemble models as a function of the data heterogeneity and speed imbalance (Section 3.1-3.3). Via a similar finite-time characterization of the FL performance, we show that the distributed estimation with network regularization outperforms FL when the number of nodes or noise variance across nodes is large (Section 3.4). We demonstrate the relative poor performance of FL when some sensors have access to highly noisy data in wireless sensor network (WSN) estimation of a Gaussian Markov random field (MRF) (Section 4.1). We also show the method's performance on a real dataset using weights proportional to the inverse of the locally estimated noise for local models (Section 4.2).


\section{Data and Processing Model}
\subsection{Data Model} \label{secsetup}

We consider a set of nodes $\mathcal{V} = \{1,\dots, N\}$ with the ability to collect and process data streams $\y_{i}=\{\y_{i,k}\in \R^m|k\in \mathbb{N}^+\}$ of the form:
\eq{y1}{\y_{i,k}=\X_i\*w^*+\ve_{i,k}+\L_i\xi_k, \quad i\in\mathcal{V}}
where $\X_i\in \R^{m\times p}$, $\*w^*\in \R^p$ is the ground truth vector of coefficients, $\{\ve_{i,k}\in\R^{m\times 1}|k \in \mathbb{N}^+\}$ are independent and identically distributed random noise variables, and $\{\xi_k\in\R^{m\times 1}|k \in \mathbb{N}^+\}$ are independent realizations of a {\em common} noise which affects nodes differently according to the matrix $\L_i\in\R^{m\times m}$. 

We assume individual noise random variables are zero-mean $\E[\ve_{i,k}]=\mathbf{0}_{m \times 1}$ and independent across different subsets, i.e., $\E[\ve_{i,k}\ve_{j,k}\trs]=\mathbf{0}_{m \times m}$ for all $i$ and $j\neq i$, and
$\E\norm{\ve_{i,k} \ve_{i,k}\trs}={ \sigma_{i}^2}\mathbf{I}_m$. 
Also, the common noise vectors are i.i.d with
$\E[\xi_k]=\mathbf{0}_{m}$ and $\E\norm{\xi_k}^2=\mathbf{I}_m$.
It follows the covariance matrix of the error term in the model for $\y_{i,k}$ as
\al{\O_{i}:=\E[(\ve_{i,k}+ \L_i\xi_k)(\ve_{i,k}+ \L_i\xi_k)\trs]={ \sigma_{i}^2}\mathbf{I}+\L_i^2 \in \R^{m\times m}.}

With $\X=[\X_1\trs, \dots, \X_N\trs]\trs$ and $\y_k=[\y_{1,k}\trs,\dots,\y_{N,k}\trs]\trs$, we can combine data streams as follows:
\eq{y_k}{\y_k=\X\*w^*+\ve_k +\L\xi_k,}
where $\L=[\L_1,\dots, \L_N]\trs$ and  $\ve_k=[\ve_{1,k}\trs, \dots, \ve_{N,k}\trs]\trs$ and the covariance matrix for the error terms is: 
\[\O:=\E[(\ve_k+ \L\xi_k)(\ve_k+ \L\xi_k)\trs]={ \Sigma}+\L\L\trs \in\R^{m\times m},   \]
where $\Sigma$ is a block-diagonal matrix with the $i$-th block equal to $\sigma^2_i \mathbf{I}_m$ and hence the noise across subsets is correlated.
A centralized formulation of the generalized least squares (GLS) consists of minimizing the following loss function over $\*w$:
\eqs{ctr}{\mathcal{L}_c \triangleq \frac{1}{2}\E[(\y_k-\X{\*w})^{\trs} \O^{-1} (\y_k-\X{\*w})].}

\subsection{A Network of ``Local" Learners}
For distributed data-processing we consider an undirected network structure $\mathcal{G}=(\mathcal{V},\mathcal{E})$ where an edge $(i,j)\in \mathcal{E}$ represents the ability to exchange information between nodes $i$ and $j$. This is also represented by the adjacency matrix
 $A \in \mathbb{R}^{N \times N}$ with $a_{i,j}=1$ if $(i,j)\in \mathcal{E}$,  and $a_{i,j}=0$ otherwise.

In a distributed and networked  estimation approach, each node $i$ solves the following ``localized" convex optimization problem,
\eq{gl2}{\min_{\*w_i}\big\{ f_i(\*w_i)+\l \rho_i(\*w)\big\},}
where
\eq{f}{f_i(\*w_i)=\frac{1}{2}\E[(\y_{i,k}-\X_i{\*w}_i)^T \O_i^{-1}(\y_{i,k}-\X_i{\*w}_i)] }
is a local loss function or measure of model fit, and $\rho_i(\*w)\geq 0$ is a measure of similarity of identified models in the neighborhood of node $i$ so that the term $\l \rho_i(\*w)$ in \eqref{gl2} can be seen as a network regularization penalty. Here we consider $L_2$ norm regularization:
\eq{pen}{\rho_i(\*w)=\frac{1}{2}\sum_{j\neq i} \hat{\alpha}_j
a_{i,j} \norm{\*w_i-\*w_j}^2,     }
where we have $\rho_i(\*w)=0$ if and only if $\*w_j=\*w_i$ for all nodes $j\neq i$ with $a_{i,j}=1$. Here $\hat{\a}_j>0$ is the weight associated to the $j$-th neighbor. We shall return to the choice of these weights in the corollaries to the main result (Theorem 2) and in our numerical illustrations. For example,
the literature on the optimal combination of forecasts (see \cite{Bates} and \cite{Granger}) suggest a choice of weights of the form $\hat{\alpha}_i = \frac{1}{\trc{\O_{i}}}$.  We note that these weights, i.e., the local data covariance matrices, are not available in a real dataset. Thus, they need to be estimated in practice, as we show in numerical examples (Section 4.2). 

The choice of network regularization parameter $\delta>0$ also plays an important role. When $\l=0$, node $i$ ignores the neighboring models and finds the model that minimizes local loss.  For large values of $\l>0$, node $i$ will favor a model closer to neighboring models, possibly at the expense of increased local loss. A similar network regularization term has been successfully used in a networked approach to multi-task learning (see \cite{Sayed_1},\cite{Sayed_2}, \cite{Sayed_4} and \cite{Sayed_5}). 

\subsection{Stochastic Gradient with Network Regularization}

In what follows, we introduce a real-time model of the computation and communication processes involved in a distributed and networked estimation approach based upon individual solutions of problem \eqref{gl2}. 
Specifically, we assume each node implements stochastic gradient updates subject to an additive network regularization penalty (SGN) based upon a noisy gradient estimate. Namely, upon collecting data point $\y_{i,k}$, node $i$ is able to compute the following gradient estimate:
\eqs{gradient}{\nabla f_{i,k}= & \X_i\trs\oi_i(\X_i\w_{i,k}-\y_{i,k})  = g_{i,k}+\X_i\trs\oi_i (\ve_{i,k}+\L_i\xi_k),}
where
$g_{i,k}:=\nabla_{\w_{i}} f_i(\w_{i,k})=\X_i\trs\oi_i\X_i(\w_{i,k}-\w^*)$.
Hence, the basic iteration of the SGN approach takes the form:
\eqs{update}{ \*w_{i,k+1}=\*w_{i,k}-\g[\nabla f_{i,k}+ \l \nabla \rho_{i,k}],}
where
$\nabla \rho_{i,k}$ is the gradient of the network regularization penalty $\rho_i(\*w_k)$ and $\g>0$ is the step size.

\begin{remark}
The basic iteration in \eqref{update} is related to the literature on consensus optimization (see e.g. \cite{nedic2009distributed}, \cite{Lian}, \cite{Yin}). However, the proposed approach can not be interpreted as being based upon {\em averaging over local} models as in consensus-based optimization. In that literature, the basic iteration is of the form:
\eqs{consensus}{
\*w_{i,k+1}=\sum_{j} \*W_{i,j}\*w_{j,k}-\gamma \nabla f_{i,k}
}
where $\*W_k \in \mathbb{R}^{N \times N}$ is doubly stochastic.
Indeed one can rewrite \eqref{update} as the form of \eqref {consensus}
with $\*W_{i,i}=1-\gamma \delta\sum_j \hat{\alpha}_j a_{i,j}$ and $\*W_{i,j}=\gamma \delta \sum_j \hat{\alpha}_{j}a_{i,j}$. However, the resulting matrix $\*W$ is \underline{not doubly stochastic} in general since we only require $\delta>0, \hat{\alpha}_j>0, j \in \mathcal{V}$. Thus, the basic iteration in \eqref{update} can not be interpreted as being based upon {\em averaging over local} models as in consensus-optimization.
\end{remark}

\subsection{Real-Time Implementation and Continuous Time Approximation}

The implementation of \eqref{update} requires that at every $k \in \mathbb{N}^+$, node $i$ has access to the current estimates of its neighbors $\{\*w_{j,k}\}_{(i,j)\in \mathcal{E}}$. Therefore to account for the real-time implementation of such updates we must model the random times required for {\em (i)} computing gradient estimates $\nabla f_{i,k}$ {\em (ii)} collect updated model parameters from neighboring nodes in order to compute the gradient of network penalty $\nabla \rho_{i,k}$.

Let $\D t_{i,k}$ be the random time required for node $i$ to compute $\tr f_{i,k}$ and $t_{i,k}=\sum_{l}^k\D t_{i,l}$ be the time $\tr f_{i,k}$ 
is obtained. We assume $\D t_{i,k}$'s  are i.i.d. with $\E[\D t_{i,k}]=\D t_i$. We define $N_{i,g}(t)=\max\{k \in \mathbb{N}^+|t_{i,k}<t\}$, and $\{N_{i,g}(t)\}$ is a renewal process. By the renewal theorem, 
\[\lim_{t\to \infty}\frac{N_{i,g}(t)}{t}=\frac{1}{\D t_i}:=\mu_i, \quad w.p. \; 1.\]

Let the time required for collecting updated models from neighbors and compute $\nabla \rho_{i,k}$  is $\D t_k$, and it holds  the same for all $i$'s. We assume $t_k$'s are i.i.d with $\E[\D t_k]=\D t$, and $t_k=\sum_l^k \D t_l$ is the time to obtain $\tr \rho_{i,k}$. Define $N_r(t)=\max\{k\in \mathbb{N}^+|t_k<t\}$, then 
\[\lim_{t\to \infty}\frac{N_{r}(t)}{t}=\frac{1}{\D t}:=\beta, \quad w.p. \; 1.\]

According to the random clock model above, each node updates its model whenever it computes the gradient of its local least squares problem ($\nabla f_{i,k}$) or receives updates from all of its neighbors. Along these lines, if node $i$ is able to process its data faster than model collection, i.e., $\mu_i>\beta$, then node $i$ would more often update its model using its local data than using the models collected from its neighbors.\footnote{The algorithm described in this section differs from the preliminary version studied in \cite{hong2020distributed} in two major ways: (i) We allow heterogeneity in processing of data across nodes, and (ii) we do not consider a specific set of averaging weights.}


A continuous-time embedding of $\w\itt$ is obtained as follows: 
\eqs{pw}{\*w_{i,t}=&\*w_{i,0} - \g \sum_{k=1}^{N_{i,g}(t)} \nabla f_{i,k}- \g
    \l \sum_{k=1}^{N_{r}(t)}\nabla \rho_{i,k} \\
     }

 Define a new variable $\w\itt:=\*w_{i,t/\gamma}$. By doing so, we ``squeeze''the timeline so that the noise terms can be approximated by Brownian motion.
In the appendix we show that the dynamics of $\w_{i,t}$  can be modeled via the stochastic differential equation (see  \eqref{A1} ):
	\eqs{dpw}{d\w\itt=&-(\mu_i g\itt+ \l\b \nabla \rho_{i,t})dt+\tau_i \X_i\trs \oi_i dB\itt + \v_i\X_{i}\trs\O_i^{-1}\L_i d B_t,}
where $\tau_i=\s_i\sqrt{ \g\mu_i}$,  and $\v_i=\sqrt{\g\mu_i}$. 
Here 
$B\itt$ and $B_t$ are the standard $m$ dimensional Brownian Motion approximating the individual noise associated with node $i$ and the common noise, respectively. In what follows, we shall characterize the convergence of the SGN scheme defined in \rf{update} via the continuous-time approximation given in \rf{dpw}.




\section{Convergence Analysis }
To characterize convergence we will use measures of {\em consistency} and {\em regularity}. 
Let $\hz$ denote a weighted average solution at time $t$, i.e.,
\eq{ww}{ \hz =\sum_{i=1}^N \alpha_i\w_{i,t}}
with {\em normalized} weights $\alpha_i:= \frac{\hat{\alpha}_i}{c} \in (0,1)$ with $c:=\sum_{i=1}^N \hat{\alpha}_i$.
Let $V\itt={\norm{\ee\itt}^2}/{2}$, where $\ee\itt:=\w\itt-\hz$. To measure regularity, we will use the weighted average difference between the solutions obtained from a single node and that of the ensemble (weighted) average: 
\eq{C}{\bar{V}_t=\sum_{i=1}^{N}\frac{\a_i\norm{\w\itt-\hat{\w}_t}^2}{2}=\sum_{i=1}^{N}\alpha_{i}V_{i,t}.}
To measure consistency we will examine the distance between the average and the ground truth,
\eq{CI}{U_t=\frac{1}{2}\norm{\hz-\tw}^2.}

\subsection{Preliminaries}
We will make use of the following definitions and results in the convergence analysis. 
 Let the Laplacian matrix of ${\mathcal{G}}$ be $L=D-A$, where $D$ is the degree matrix, and $A$ is the adjacency matrix. 
We define the generalized Laplacian matrix as $\hat{L}=\hat{D}-\hat{A}$, where $\hat{D}$ is a diagonal matrix whose $i$-th diagonal entry is equal to $\sum_{j\in \mathcal{V}}\hat{\alpha}_{i}\hat{\alpha}_{j}a_{i,j}$, and $\hat{A}$ is the weighted adjacent matrix with $A_{i,j}=\hat{\a}_i\hat{\a}_ja_{i,j}$.  Let $\lambda_2 $ (respectively, $\hat{\lambda}_2$) denote the second smallest eigenvalue of $L$ (respectively, $\hat{L}$).


The continuous-time gradient $g\itt$ defined above is a function of $\w\itt$, which we do not explicitly specify to simplify notation.
In our analyses, we denote
$g\itt(\hz)=\X_i\trs\oi_i\X_i(\hz-\w^*)$ and
$g\itt(\w^*)=\X_i\trs\oi_i\X_i(\w^*-\w^*)$.  Note that $g\itt(\w^*)=0$ for  all $i\in \mathcal{V}$ and $t$, to simplify notation we will write $g(\w^*)$ instead.  Similarly, when a property holds for all $t$, we drop $t$ and write  $g\itt$ as $g_i$.  

We note  that $g_i$'s are $\eta$-Lipschitz continuous and the corresponding loss function (noise-free version of $f_i$) is strongly convex with $\k_i$. 
Let $\w_{i,1}$ and $\w_{i,2}$ be two input vectors taken from the function domain, then 
\eqs{lip}{\norm{ g_i(\w_{i,1})- g_i(\w_{i,2}  )}& =\norm{\X_i\trs\oi_i\X_i(\w_{i,1}-\w_{i,2})}\\
	& \leq \eta_i \norm{\w_{i,1}-\w_{i,2}} \leq \eta \norm{\w_{i,1}-\w_{i,2}},
}
where $\norm{\cdot}_F$ is the Frobenius norm and
$\eta_i:= \norm{\X_i\trs\oi_i\X_i}$, $\eta:=\max_i\norm{\X_i\trs\oi_i\X_i}$. Furthermore,  
\eqs{conv}{( g_i(\w_{i,1})- g_i(\w_{i,2}))\trs(\w_{i,1}-\w_{i,2})
&\geq \k_i\norm{\w_{i,1}-\w_{i,2}}^2 \geq \k\norm{\w_{i,1}-\w_{i,2}}^2,
}
for  $\k_i:= 2\lambda_{min}(\X_i\trs\oi_i\X_i)$ and $\k:=\min \k_i$, where $\lambda_{min}(\cdot)$ is the smallest  eigenvalue. By Definitions \rf{strong} and \rf{strict}, $g_i$ is strongly convex with $\kappa_i$ and the corresponding loss function is Lipschitz continuous gradients with constant $\eta_i$.

The ratio $\frac{\eta_i}{\kappa_i}>1$ is referred to as the condition number of the deterministic optimization problem $\min_{\w} f_i(\w)$. In the analysis below, the {\em worst-case} condition number, i.e. $\frac{\eta}{\kappa}>1$, plays an important role in characterizing performance.


%
%
%

\subsection{Regularity}
The following result establishes that the sum of regularity and consistency measures is equivalent to the weighted sum of the distances between individual solutions $\w\itt$ and the ground truth $ \w^*$.
\begin{lemma} \label{tri} 
	Consider $\bar{V}_t$ in \eqref{C} and $U_t$ in  \eqref{CI}. We have 
	\eq{d}{\frac{1}{2} \sum_{i=1}^N \a_i\norm{\w\itt-\tw}^2
	= \bar{V}_t+U_t.}
\end{lemma}

\proof 
We expand the sum on the left-hand side of \eqref{d} as follows,
\eqs{de}{\frac{1}{2}\sum_{i=1}^N\a_i\norm{\w\itt-\tw}^2 &= \frac{1}{2}\sum_{i=1}^N\a_i\Big[\norm{\w\itt-\hz}^2+\norm{\hz-\tw}^2+ 2 \vb{\ee\itt,\hz-\tw}   \Big].}
Note that
\eq{s0}{ \sumbs{\ee\itt}=\sumbs{\w\itt}-\hz=0.}
Hence, \eqref{d} follows by observing that the summation of the cross-product (last) term inside the brackets in \eqref{de} is zero.  
\endproof

In what follows, we obtain upper bounds on the expectations of regularity $\bar{V}_t$ and consistency $U_t$ processes in Theorems \ref{v} and \ref{u}, respectively. Given the relation in Lemma \ref{tri}, these bounds provide a bound on the average error of individual estimates generated by the SGN algorithm with respect to the ground truth. 
The following result provides an upper bound on the expected regularity of the estimates at a given time.
\begin{theorem}\label{v} Let $\w\itt$ evolve according to continuous time dynamics \eqref{dpw}. Then
\al{\E[\bar{V}_t]& \leq e^{-2(\mu\k+\l\b\hl) t}\bar{V}_0+ \frac{ C_1}{2(\mu\k+\l\b\hl)}( 1-e^{-2(\mu\k+\l\b\hl) t}),}
	where $C_1>0$  is defined as follows:
		\eqs{c1}{C_1:=\frac{1}{2}\sum_{i=1}^N\a_i C_{1,i}, }
		with $C_{1,i}$ for $i\in \mathcal{V}$ defined as, 
		\eqs{cts}{C_{1,i}&:=	  \tau_i^2 \Big( 1-2\a_i  \Big)\norm{\X_i\trs\oi_i}^2_F+ \sum_{k=1}^N\a_k^2\tau_k^2 \norm{ \X_k\trs\oi_k  }_F^2 \\
		& ~~~~~~~+\v_i^2\norm{\X_i\oi_i\L_i}_F^2+\sum_{k=1}^N\sum_{j=1}^{N}\a_k\a_j\v_k\v_j A_{k,j} -2 \sum_{k=1}^N\a_k\v_i\v_k A_{i,k},   }
with $A_{i,k}:=\mathbf{1}\trs( \X_i\trs\oi_i\L_i \circ \X_k\trs\oi_k \L_k) \mathbf{1}$ and ``$\circ$" denoting the Hadamard product. 
	In the long run, \[\lim_{t\to\infty}\E[\bar{V}_t]\leq \frac{ C_1}{2(\mu\k+\l\b\hl)}.\]
\end{theorem}
\proof
See Appendix \ref{C1}.
\endproof

It is not surprising that the expected difference in estimates decreases with growing $\l$, which penalizes disagreement with neighbors. Similarly, the larger the algebraic connectivity of the network $\hl$ or the strong convexity constant $\kappa$ is, the smaller is the expected $\bar V_t$. 

Finally, the constant term $\C$ (defined in \eqref{c1}-\eqref{cts}) is determined by $\X_i, i \in \mathcal{V}$ and the covariance matrices of the common noise $\L_i,i \in \mathcal{V}$.
 As we show in the appendix, the quadratic variation of $\ee\itt$ is: 
\[ \vb{\ee\itt}_{t}=\E\int_0^{t} \frac{1}{2}C_{1,i} ds\to \var(\ee\itt).\]
Hence, $\frac{t}{2}C\itt$ describes the variation of $\ee\itt$, and  $C_1$ is a weighted measure of variation.
$\C$ is small when we have nodes that are less affected by the noise. 

\subsection{Consistency}

The consistency measure $\{U_t, t\geq 0\}$ captures the performance of the average solution $\hat \w$. In the following theorem, we provide a characterization of the performance of the collective effort.

\begin{theorem}\label{u} Let $\w\itt$ evolve according to continuous time dynamics \eqref{pw}. Then
\al{ \E[U_t]\leq & e^{-2\mu\k t}U_0 + \frac{1}{2\k\mu}\Big(\frac{\eta\mu'-\k\mu}{\l\beta\hl  }C_1+C_2 \Big)\Big(1-e^{-2\mu\k t}\Big),}
	where $C_1$ is defined in \rf{c1}, $\mu'=\max_i \mu$, and $\mu=\min_i \mu_i$, and 
 
	\eqs{c2}{C_2&=\frac{1}{2}\Big(\sum_{k=1}^{N}\a_k^2\tau_k^2\norm{\X_k\trs\oi_k}_F^2+\sum_{k=1}^N\sum_{j=1}^{N}\a_k\a_j\v_k\v_jA_{k,j} \Big).}
	In the long run, 
\eqs{eu}{\lim_{t\to\infty}\E[U_t]\leq & \frac{1}{2\mu\k}\Big(\frac{\eta\mu'-\k\mu}{\l\beta\hl  }C_1+C_2\Big).}	

\end{theorem}

\proof
See Appendix \ref{D1}. 
\endproof

Similar to the regularity measure bound, the penalty constant $\l>0$ and the algebraic connectivity $\hl$ reduce the bound on the expected consistency. However, the long-run expected difference between the ensemble average estimate and the ground truth does not reduce to zero as $\l \hl \to \infty$. The constant $\js$, determined by the data $\X_i$ and matrices $\L_i$, captures the performance gap in the long run due to available data.  Note that the quadratic variation of $\hz-\w^*$ is as follows, 
\[\vb{\hz-\w^*}_{t}=\E\int_0^{t} C_2 ds\to \var(\hz-\w^*).\]
Hence $tC_2$ describes the variation of $\hz-\w^*$. In the following result, we examine asymptotic performance as the network grows in size.


\begin{lemma}\label{w} (Asymptotic Performance)
	Assume the network of local learners (with associated data sets) grows as follows:
	\begin{itemize}
		\item 
		Each new node (say $n>N$, $N=1,2,\dots$) is associated with a new dataset $\X_n$ and $\y_{n,k}$ given by \rf{y1} with  
		\begin{align*}
			\|X_n\|_F<L_0, & ~~~\trc{\O_{n}}\leq { M}, \text{ and }   \e\leq\s_n^2,
		\end{align*} where $L_0<\infty$, and $ 0<\e<M<\infty$. We assume all entries of the weight matrix $\L_i$'s are bounded above by $\o_2$, the maximum entry of all weight matrices. 
		\item The network connectivity is preserved.
	\end{itemize}
	If we have $\l \hl \sim N$, and $\g\sim \e^3$, then 
	 
	\al{\lim_{t\rightarrow \infty}\lim_{N\rightarrow \infty}\E[\norm{\hz-\tw}^2]=&\lim_{t\rightarrow \infty}\lim_{N\rightarrow \infty}2\E[U_t] <\frac{S_2\mu'\g m \, \o_2}{\k\mu\e^2}=O(\g^{1/3}),  } 
where $S_2=\max_{k,j}\max_{i}\x_{i,k}\trs\x_{i,j}$ with $\x_{i,k}$ being the $k$-th column of $\X_i\trs$. 
\end{lemma}


The proof (see Appendix \ref{F1}) follows by constructing an upper bound for \rf{eu} in Theorem \ref{u} by considering constants $\C$ and $\js$. We first show that all terms of $\js$ can be bounded by $\frac{S_2\mu'\g m \o_2}{\e^2}$ by using the assumptions on each new dataset available to each new node. This bound on $\js$ increases with the noise term that affects the nodes,  inputs magnitude,  and step size. Second, show that $\C/N$ goes to zero as $N\to \infty$. Combining the limiting properties of the two constants with the bound in \eqref{eu} and selecting $\l$ such that $\l \hl\sim N$, we obtain the bound above. This bound can be controlled by the selection of the step size $\g$. Note that we can make $\l \hl \sim N$ by adjusting the penalty parameter when the network retains connectivity as the number of nodes grows. 

\begin{remark}
	Lemma \ref{w} provides an upper bound  of the difference between the ensemble average estimate and the ground truth, which is $O(\g^{1/3})$ as $N \rightarrow \infty$.  The bound is subject to the influence of the data:  smaller in  magnitude (smaller $S_2$) leads to a tighter bound.
	We observe that the bound is tighter when the objective function is smoother (larger $\k$).
	Since we require $\g\sim \e^3$ ( or even smaller $\g$), the effect of having small $\e$ is offset  by choosing smaller $\g$. 
	Thus the bound can be controlled as small as needed by choosing the stepsize $\g$.
\end{remark}

\subsection{Comparison with Federated Learning (FL)}
In this section,  we firstly present the convergence property of FL, and then  compare the performance of SGN with the FL. 

\subsubsection{Federated Learning}
For the federated learning, the gradient samples are sent to a central server for computing, we define the FL approximation at $t$ as:
\[ \w_t=\sum_i\w\itt,\]
and the continuous time embedding of FL updates is as follows:
  \eqs{fl1}{\*w_{t}=\*w_{i,0} - 
 	\g \sum_{i=1}^N\sum_{k=1}^{N_{i,g}(t)} \nabla f_{i,k}+\sum_{i=1}^N\sum_{k=1}^{N_{i,g}(t)} \e_{i,k}}
Define a new variable $\w_t:=\*w_{t/r}$, then the corresponding  continuous time dynamics of $\w_t$ can be approximated by:
\eqs{fl2}{d\w_t=&\sum_{i=1}^{N}\Big[-\mu_i g\itt  dt +\tau_i \X_i\trs \oi_i  dB\itt +\v_i\X_{i}\trs\O_i^{-1}\L_i d B_t\Big].  }
Now we define the measure $F_t$ to examine the distance between the FL estimates and the ground truth and present the  bounds on the expectation of $F_t$ in the following theorem.

\[F_t=\frac{1}{2}\norm{\w_t-\tw}^2.\]

\begin{theorem}\label{FL} Let $\w_t$ evolve according to continuous time dynamics \eqref{fl1}, then
	\al{dF_t=&  -\sum_{i=1}^{N}\mu_ig\itt\trs(\w_t-\tw) dt+K_3d\td{B}_{f,t}+C_3 dt,   }
where $K_3d\td{B}_{f,t}$ is the summation of the Ito terms, 
\eqs{kk3}{K_3d\td{B}_{f,t}=& \sum_{i=1}^{N}\tau_i(\w_t-\tw)\trs \X_i\trs \oi_idB\itt + \sum_{i=1}^{N}\v_i(\w_t-\tw)\trs\X_{i}\trs\O_i^{-1}\L_i d B_t,}
and $C_3$ is the summation of the constant terms, 
\eqs{cc3}{ C_3=&\frac{1}{2} \Big(  \sum_{i=1}^{N}\tau_i^2\norm{ \X_i\trs \oi_i }_F^2+\sum_{k=1}^N\sum_{j=1}^{N}\v_k\v_j A_{k,j}   \Big).}
Moreover,
	\[\E[F_t]\geq e^{-2\eta \mu' t}F_0+\frac{1}{2\eta \mu'}  C_3(1-e^{-2\eta \mu' t}),\]
	\[\E[F_t]\leq e^{-2\k \mu  t}F_0+\frac{1}{2\k\mu}  C_3(1-e^{-2\k \mu t}),\]
so that in the long run, 
	\al{\frac{C_3}{2\eta\mu'} \leq \lim_{t\to\infty}\E[F_t]\leq & \frac{C_3}{2\k\mu} .}
\end{theorem}
\proof See Appendix \ref{T3}.
\endproof

In this theorem, we provide both upper and lower bound of $F_t$. The strong convexity parameter $\k$ and the Lipschitz constant $\eta$ influence the bounds.
The constant $C_3$, determined by the data $\X_i$ and matrices $\L_i$ determines the quality of the asymptotic estimator derived by federated learning, and   $tC_3$ describes the variation of $\w_t-\w^*$.

\subsubsection{Comparison with Federated Learning}

In this section, we compare the long run performance of the SGN scheme and FL scheme with streaming data. Specifically, we will compare the upper bound in Theorem 2 and the lower bound in Theorem 3. This is tantamount to comparing  $\frac{C_2}{\mu \kappa}$ (with $C_2$ defined in \rf{c2}) and $\frac{C_3}{\mu' \eta}$ (with $C_3$ defined in \rf{cc3}) for large enough values of $\delta>0$.

\begin{corollary}\label{FLNR}
Assume $\hat{\alpha}_{i}=1, i \in \mathcal{V}$, so that $\alpha_i=\frac{1}{N}$, (i.e. simple
average) and $N>\sqrt{\frac{\mu ^{\prime }\eta }{\mu \kappa }}$. There exists $\bar{\delta}<\infty$ such that for all $\delta>\bar{\delta}$ it holds that:
\[
\lim_{t\rightarrow \infty }E[U_{t}]<\lim_{t\rightarrow \infty }E[F_{t}].
\]
\end{corollary}
\proof
If the weights are of the form $\alpha _{i}=\frac{1}{N}$ (i.e. simple
average) then%
\begin{eqnarray*}
C_{2} &=&\frac{1}{2}\left( \sum_{k=1}^{N}\alpha _{k}^{2}\tau
_{k}^{2}\left\Vert \mathbf{X}_{k}\trs\boldsymbol{\Omega }_{k}^{-1}\right\Vert
_{F}^{2}+\sum_{k=1}^{N}\sum_{j=1}^{N}\alpha _{k}\alpha _{j}\varsigma
_{k}\varsigma _{j}A_{k,j}\right)  \\
&=&\frac{1}{2N^{2}}\left( \sum_{k=1}^{N}\tau _{k}^{2}\left\Vert 
\mathbf{X}_{k}\trs\boldsymbol{\Omega }_{k}^{-1}\right\Vert
_{F}^{2}+\sum_{k=1}^{N}\sum_{j=1}^{N}\varsigma _{k}\varsigma _{j} A_{k,j}\right)  \\
&=&\frac{C_{3}}{N^{2}}.
\end{eqnarray*}
It follows that for $N>\sqrt{\frac{\mu ^{\prime }\eta }{\mu \kappa }}$, we have $\frac{C_{2}}{\mu \kappa }<\frac{C_{3}}{\mu ^{\prime }\eta }$, and the lower bound in Theorem 3 exceeds the
the upper bound in Theorem 2 whenever:
\begin{equation}
\label{inequality}
\frac{C_{2}}{\mu \kappa }+\frac{1}{\mu \kappa }(\frac{\mu ^{\prime }\eta
-\mu \kappa }{\delta \beta \lambda _{2}})<\frac{C_{3}}{\mu ^{\prime }\eta, }
\end{equation}
or equivalently,%
\[
\frac{1}{\delta \beta \lambda _{2}}(\frac{\mu ^{\prime }\eta }{\mu \kappa }%
-1)<\frac{C_{3}}{\mu ^{\prime }\eta }-\frac{C_{2}}{\mu \kappa }.
\]%
Hence, for $\delta >\bar{\delta}$ with 
\[
\bar{\delta}:=\frac{1}{\beta \lambda _{2}}(\frac{\mu ^{\prime }\eta }{\mu
\kappa }-1)(\frac{C_{3}}{\mu ^{\prime }\eta }-\frac{C_{2}}{\mu \kappa })^{-1}.
\]
The inequality \eqref{inequality} holds and the result follows.
\endproof

The above result guarantees that a large enough network ensures an ensemble averages estimate with higher quality than that obtained with federated learning. 
In the following corollary, we show that even for small networks but highly heterogeneous datasets (high asymmetries in individual noise), the SGN ensemble estimate is superior to the estimate obtained through federated learning.

\begin{corollary}
Consider the case with $\boldsymbol{\Lambda }_{i}=0$ for all $i\in \mathcal{V%
}$ and
\[
\begin{array}{rrr}
\hat{\alpha}_{i}=\frac{1}{\sigma _{i}^{2}}, &  & c=\sum\nolimits_{k=1}^{N}\frac{1%
}{\sigma _{k}^{2}},%
\end{array}%
\]%
let $\bar{\sigma}>0$ be defined as:%
\[
\frac{1}{\bar{\sigma}^{4}}:=\frac{\sum_{k=1}^{N}\frac{1}{\sigma
_{k}^{4}}\mu_k\left\Vert \mathbf{X}_{k} \right\Vert _{F}^{2}}{%
\sum_{k=1}^{N}\mu _{k}\left\Vert \mathbf{X}_{k}\right\Vert _{F}^{2}}.
\]%
If $\frac{\bar{\sigma}^{2}}{\min \sigma _{k}^{2}}>\sqrt{\frac{\mu ^{\prime
}\eta }{\mu \kappa }}$, there exists $\bar{\delta}<\infty $, such that for all 
$\delta >\bar{\delta}$, it holds that%
\[
\lim_{t\rightarrow \infty }\E[U_{t}]<\lim_{t\rightarrow \infty }\E[F_{t}].
\]

\end{corollary}

\proof 
In this case, $\boldsymbol{\Omega }_{k}^{-1}=\frac{1}{\sigma _{k}^{2}}%
\boldsymbol{I}_{k}$, therefore $\left\Vert \mathbf{X}_{k}\trs\boldsymbol{%
\Omega }_{k}^{-1}\right\Vert _{F}^{2}=\frac{1}{\sigma _{k}^{2}}\left\Vert 
\mathbf{X}_{k}\right\Vert _{F}^{2}$. Also, $\alpha_i =\frac{\hat{\alpha}_i}{c}$ and
\[
C_{2}=\frac{\gamma }{2}\sum_{k=1}^{N}\alpha _{k}^{2}\sigma _{k}^{2}\mu
_{k}\left\Vert \mathbf{X}_{k}\trs\boldsymbol{\Omega }_{k}^{-1}\right%
\Vert _{F}^{2}=\frac{\gamma }{2c^{2}}\sum_{k=1}^{N}\frac{\mu _{k}}{\sigma
_{k}^{4}}\left\Vert \mathbf{X}_{k}\right\Vert _{F}^{2},
\]
and%
\[
C_{3}=\frac{\gamma }{2}\sum_{k=1}^{N}\sigma _{k}^{2}\mu _{k}\left\Vert 
\mathbf{X}_{k}\trs\boldsymbol{\Omega }_{k}^{-1}\right\Vert _{F}^{2}=%
\frac{\gamma }{2}\sum_{k=1}^{N}\mu _{k}\left\Vert \mathbf{X}_{k}\right\Vert _{F}^{2}.
\]%
Hence, 
\[
\frac{C_{3}}{C_{2}}=c^{2}\bar{\sigma}^{4}.
\]%
Since $\frac{1}{c}\leq \min_{k}\sigma _{k}^{2}$, we conclude that if $\frac{\bar{\sigma}%
^{2}}{\min_{k}\sigma _{k}^{2}}>\sqrt{\frac{\mu \kappa }{\mu ^{\prime }\eta }}
$, then $\frac{C_{2}}{\mu \kappa }<\frac{C_{3}}{\mu ^{\prime }\eta }$. The rest of the proof follows the same argument to the previous corollary.
\endproof

This second corollary indicates that with large differences in individual noise variance, the SGN approach provides a higher quality asymptotic estimate than federated learning.
	


\section{Numerical Illustrations}

In this section, we apply the proposed method to three examples to corroborate the analytical results. First, we apply the SGN algorithm to a MRF estimation problem using a WSN with synthetic data and compare our scheme with FL. Next, we look at a real-world problem: the escape behavior of European gregarious birds.  Finally, we consider a synthetic dataset to analyze the effects of the number of nodes and network structure. 



\subsection{Temperature Estimation of a Field}\label{ex2}
We consider a WSN deployed to estimate the temperature on a 10m$\times$10m field, divided into $100$ equal squares. We assume that the temperature within the same square is the same, and the field's true temperature is stored in the vector $\tw\in\R^{100\times 1}$. We randomly place $N$ sensors on the field, and each measures $\tw$ using noisy local observations $\y_i\in \R^{100\times 1}$, which is corrupted by the measurement noise $\ve_i$, a detection error that only influence sensor $i$; and the network disturbance $\xi$, a common noise that is shared by all sensors. Each sensor $i$ only shares a portion of $\xi$, which is determined by a matrix $\L_i$. $\L_i$ values reflect the relative distance between the sensors' locations and the measured squares: a sensor that is close to a square is subject to lower noise levels. We assume $\tw $ is fixed but $\y_i$ changes at each measurement, which can be expressed as follows:
\eqs{ex_y}{\y_i=\tw+\ve_i+\L_i\xi,}
where  $\ve_i\sim \mathcal{N}_m(\mathbf{0}, \s^2 \mathbf{I})$ and $\xi\sim \mathcal{N}_m(\mathbf{0},  \mathbf{I})$. Note that if we set $\X_{i}=\mathbf{I}$,  \rf{y1} and  \rf{ex_y} have the same form. 

We model the temperature of the field using a Gaussian MRF and let the temperature values range from $\ang{0}$F to $\ang{255}$F, as in \cite{eksin2012distributed}. Two heat locations are located at $(2m, 8.5m)$ and $(8.5m, 9m)$, and temperature drops from the heat source at a rate of $\ang{25}$ F/m within an area of influence of $5m$ from the source. We set $N=40$ and randomly connect the nodes to their neighbors within $2.5$ $m$---see Figure 1(a) for sensor locations and heat map of the field.
Each node has the following local cost, 
\eqs{cost}{\mathcal{L}(\w_i) =&\frac{1}{2}(\y_i-\w_i)\trs\oi_i(\y_i-\w_i)
+\frac{\delta}{2}\sum_{j\in \mathcal{V}\setminus i}   a\ij\hat{\alpha}_j\norm{\w_i-\w_j}^2.}
In this experiment, we set $\hat{\alpha}_i=\frac{1}{\trc{\O_i}}$ for each $i$. 

		
	

\begin{figure}[htp]
	\centering
	\subfloat[]{\label{figur:1}\includegraphics[width=55mm]{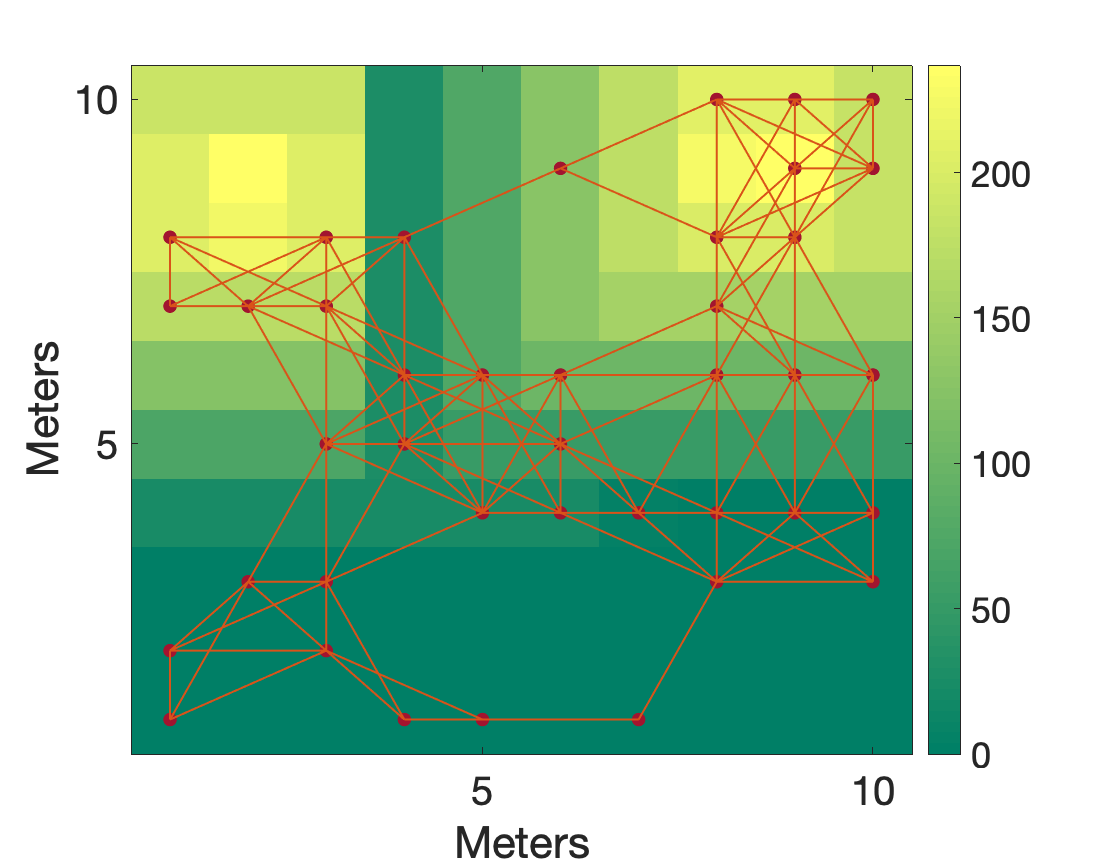}}
	\subfloat[]{\label{figur:2}\includegraphics[width=55mm]{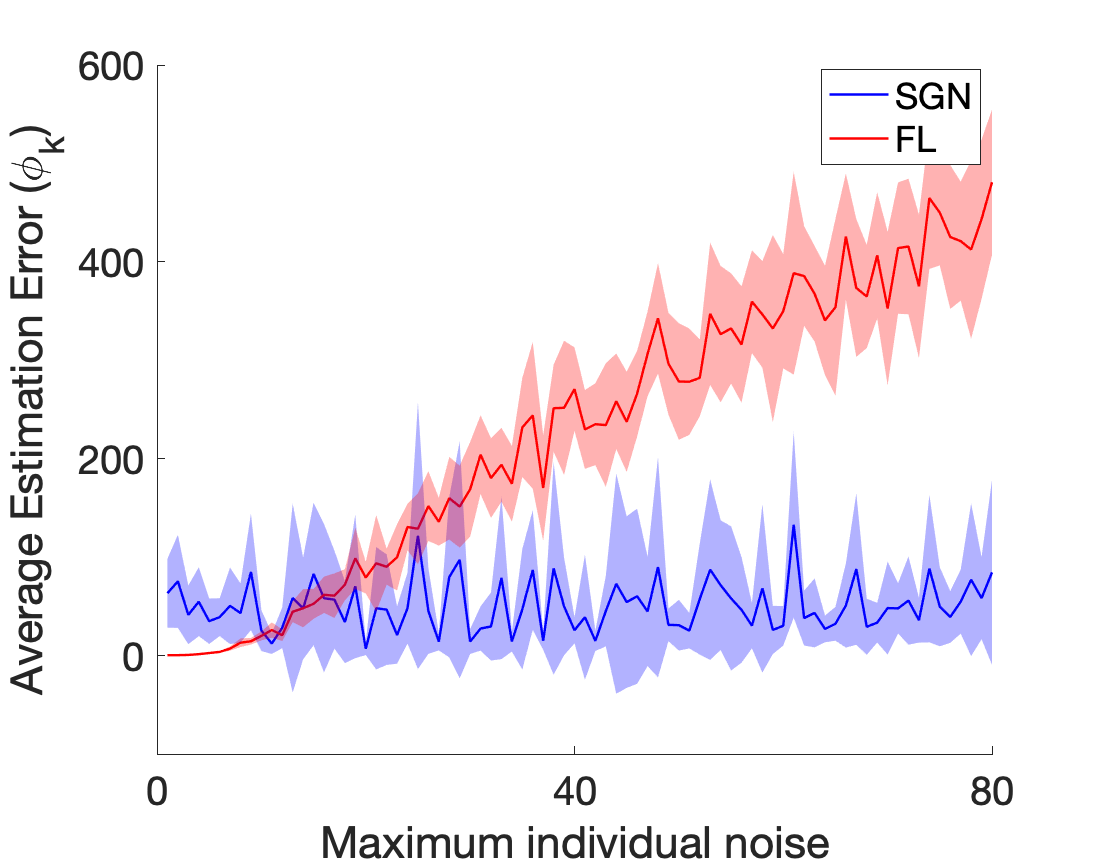}}\\
	\begin{minipage}{9cm}%
		\small Figure 1. Sensor network and SGN and FL estimation results. (a) Network structure of sensors. The orange dots denote the nodes, and the  lines represent the edges between nodes. The two heat sources are located at $(2m,8.5m)$ and $(8.5m, 9m)$, marked by yellow. (b)  The $95\%$ confidence intervals of average estimation error $\bar{\phi}_k$ by SGN and FL at the final iteration.
		
	\end{minipage}%
	
\end{figure}

\begin{figure}[htp]
	\centering	
	\subfloat[]{\label{figur:1}\includegraphics[width=40mm]{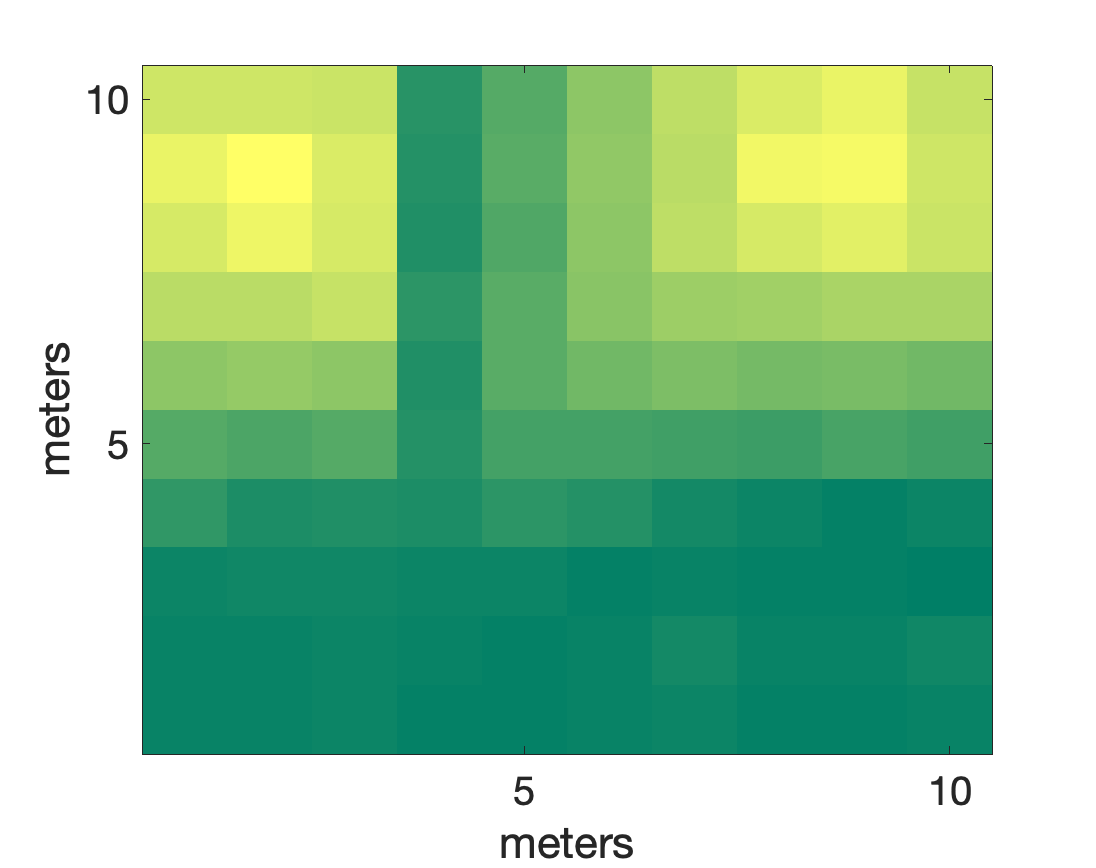}}
	\subfloat[]{\label{figur:2}\includegraphics[width=40mm]{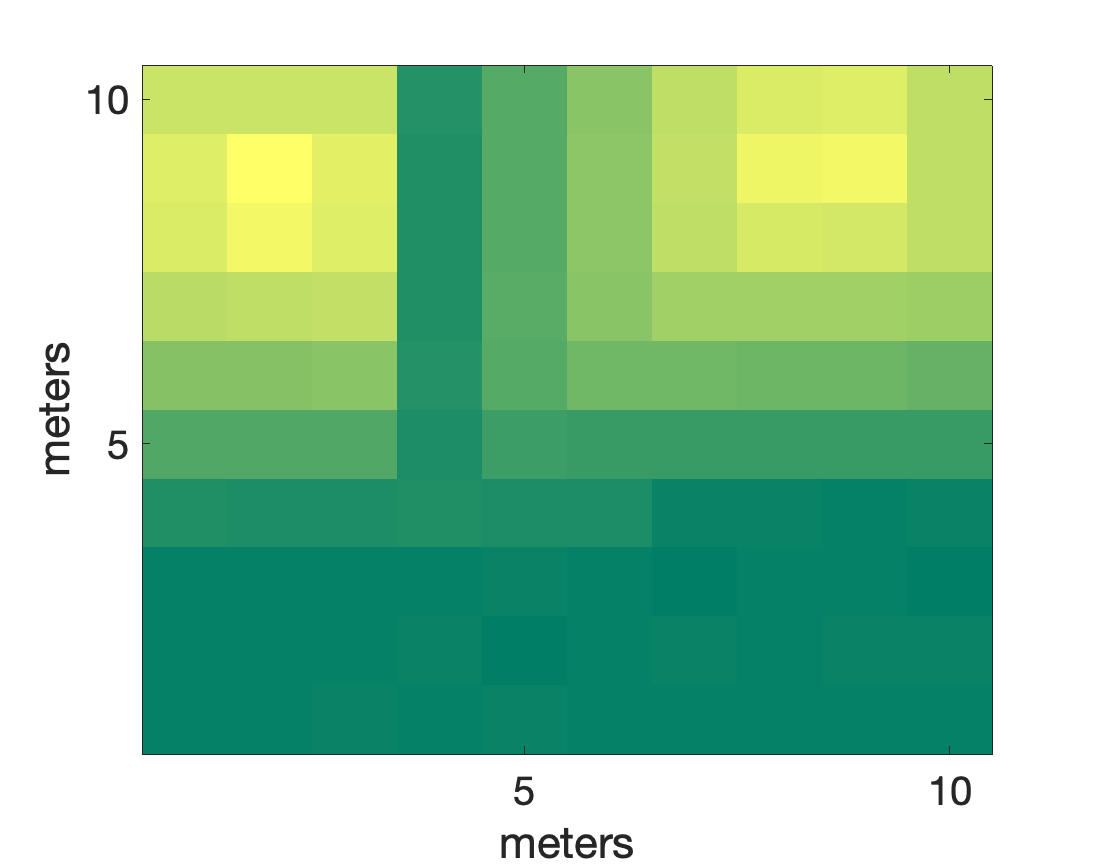}}
	\\
	\subfloat[]{\label{figur:3}\includegraphics[width=40mm]{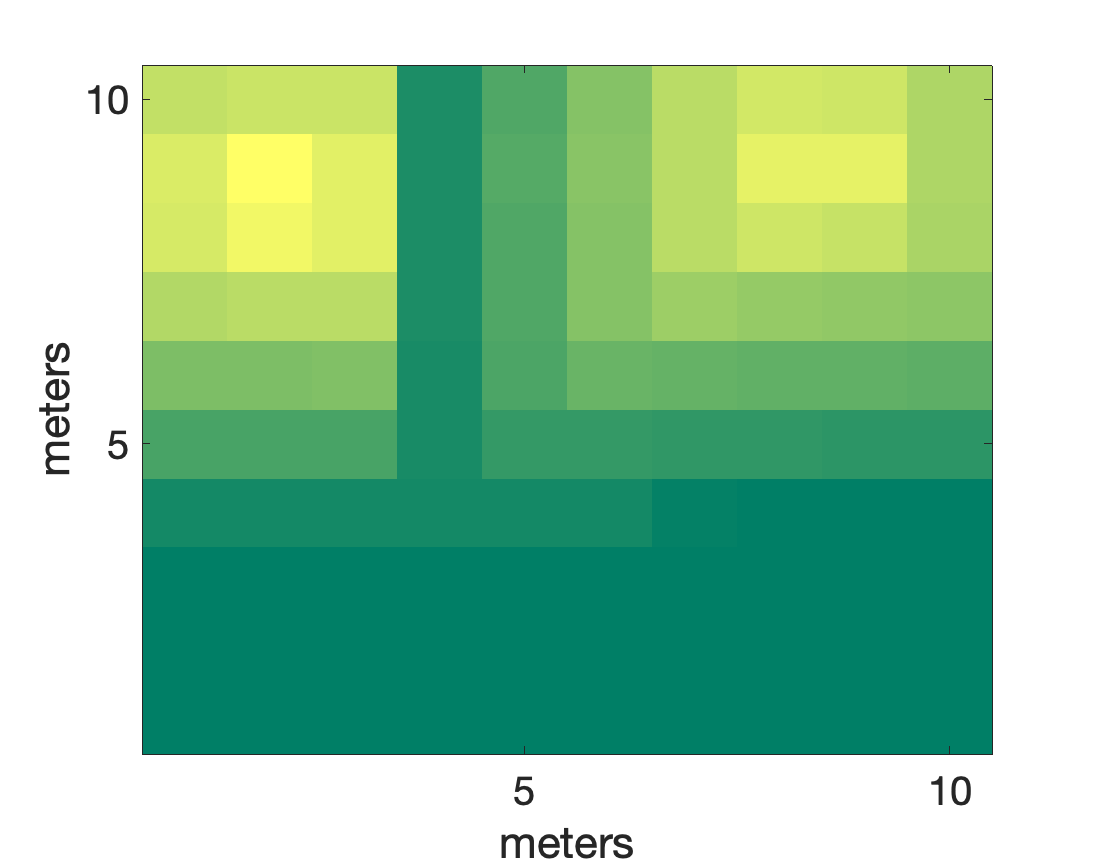}}
	\subfloat[]{\label{figur:4}\includegraphics[width=40mm]{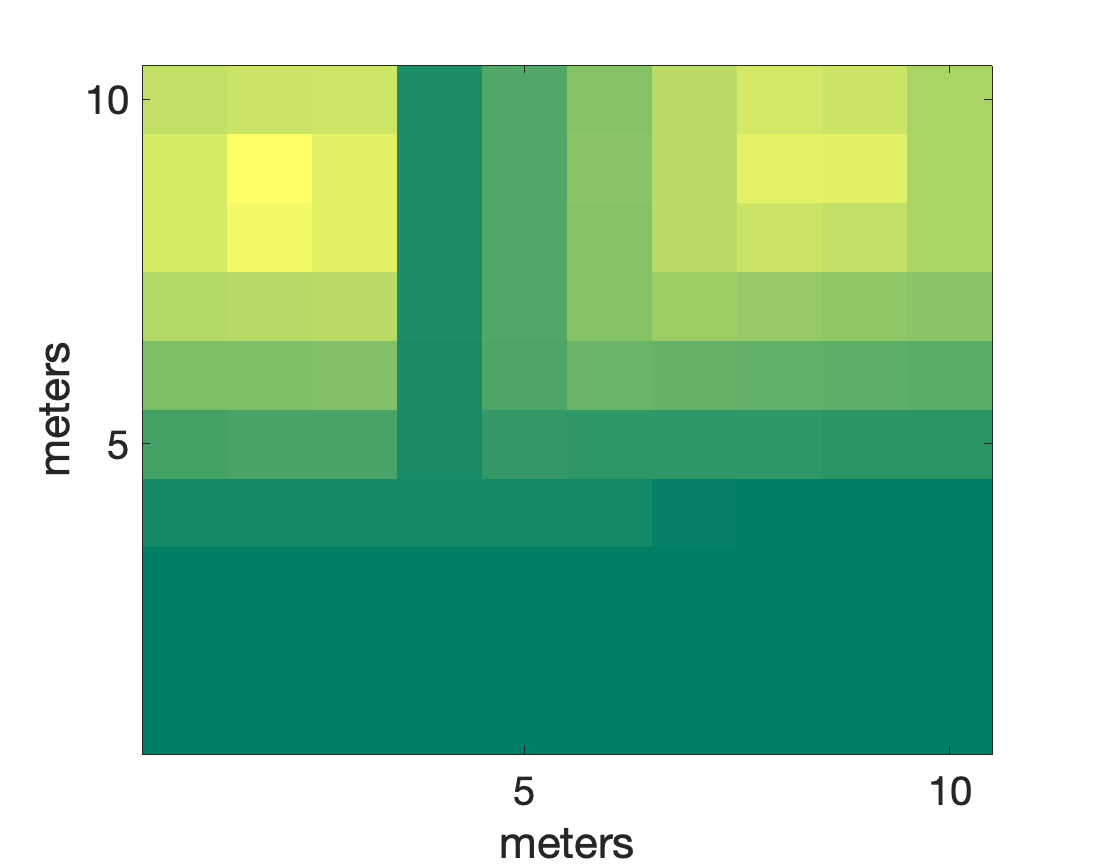}}
	\\
	\begin{minipage}{9cm}%
		\small Figure 2. Field temperature estimation from a single node (the $1$st sensor). (a)-(d) display the sensor estimations at time points $k=2$, $k=4$, $k=20$, and $=100$, respectively. 
	\end{minipage}%
	
\end{figure}

We would like minimize the cost function \rf{cost} using SGN and FL for each sensor by selecting proper $\w_i$. We set the stepsize $\g=10^{-5}N$ for SGN and $\g=10^{-5}$ for FL, the penalty parameter is set as $\l=100$ for SGN. We set the minimum individual noise variance as $0.01$ and let the maximum change from $1$ to $80$. The stream parameters are set as $\mu_i=10\s_i^2$, i.e., the faster nodes also generate noiser data. 
We define the estimation error at time point $k$ from the $l$th trial as \eqs{phi}{\phi_{k,l}=\norm{\hat{\w}_k^{(l)}-\w^*},}
where $\hat{\w}_{k}^{(l)}$ is the weighted average \eqref{ww} from the $l$th trial. The average estimation error over $K$ trials is defined as:
\eqs{phib}{\bar{\phi}_k=\frac{1}{K}\sum_{l=1}^K\phi_{k,l}.}
We run both algorithms for $10$ trials and show the average estimation error  with $95\%$ confidence interval at iteration $T=4000$ in
Figure 1(b). It shows at the final iteration, as the maximum individual noise variance increases, the average estimation error of FL increases while that of SGN is stable, which suggests that SGN is robust against noise.

Figure 2 presents the estimations of a single node (the $1$st sensor).
The estimates are noisy at the early stage ($k=2$). By the time $k=5$, we observe significant noise reduction compared to the first few iterations. This reduction becomes more substantial at $k=20$. At  $k=100$, the temperature estimations of $1$st sensor are close to the true temperature values.

\subsection{ Modeling Flocking Escape Behavior}\label{ex3}

In this section, we consider a real data-set used for modeling bird escape behavior based upon the Flight Initiation Distance (FID), the distance at which animals take flight from approaching threats (see \cite{blumstein}). In a regression model, the FID is considered as the response variable, and the predictors are {\em flock\_size} (the number of aggregated individuals of the same species), {\em start\_dist} (the distance at which a predator started the approach to the bird), {\em habitat} (a binary variable with urban$=1$ and rural$=0$), {\em latitude} (of the study location), {\em diet} (primary type of food the bird consumed, all species were classified into $3$ main categories: granivorous(g), granivorous–insectivorous(gi), and insectivorous(i))\cite{morelli2019}. We transform the variable ``diet" into $3$ binary variables, each indicating one diet category, and normalize all other variables for the following analysis. At each node, the FID estimate is given by %
\al{\widehat{FID}=&w_0+w_1\cdot\text{start\_dist}+w_2\cdot\text{diet(gi)}+w_3\cdot \text{diet(g)}\\&+w_4\cdot\text{diet(i)}+w_5\cdot\text{latitude}+w_6\cdot\text{flock\_size}+w_7\cdot\text{habitat},}
where we denote node $i$'s estimate with $\w_i=[w_0,\dots, w_7]$ as before. The data contains $941$ observations in total collected from eight European countries and 23 different bird species.

We group $23$ bird species into $N= 15$ nodes where each species is assigned to one node. 
Since all nodes contain more than $10$ observations, we use the mini-batch of size $10$ in the experiment to unify the data length at different nodes. In this experiment, we consider a  complete network (a network with lower connectivity may only reduce the performance slightly). We use $\hat{\alpha}_i=\frac{1}{\trc{\O_i}}$ for each $i$, and the covariance matrix of a node is computed as the diagonals of the covariance matrix of $15$ mini-batch samples. We use a fading memory update rule to compute the trace of the covariance matrix, $\trc{\O_i}$, \cite{nocedal}:
\al{\trc{\O_{i,k+1}}=\varphi\trc{\O_{i,k}}+(1-\varphi)\trc{\hat{\O}_{i,k+1}},   } 
where $\trc{\hat{\O}_{i,k+1}} $ is the $i$th covariance matrix trace computed at the $(k+1)$th iteration, and $\varphi \in (0,1)$ is the fading parameter that controls the memory of the past covariance values.  For SGN experiments in this section, we set parameter $\varphi=0.9$, the step size $\g=0.001$, and the regularization penalty $\l=100$.

The SGN's estimation error at time point $k$ from the $l$th run                                is given by
$\phi_{k,l}=\norm{FID-\X\hat{\w}_k^{(l)}}$,
where $\X$ is the matrix containing the observations and $\hat{\w}_k^{(l)}$ is the SGN weighted estimation \eqref{ww} from the $l$th trial. The average estimation error $\bar{\phi}_k$ is defined as in \eqref{phib} with $K=20$. Figure 3(b) shows that the average estimation error $\bar \phi_k$ of SGN converges after 2000 iterations. 
We compare the final estimator $\hat \w_T$ at $T=3000$ of  SGN with the solution of the GLS in Table 1. Half of the SGN estimators fall into the 97\% confidence interval(CI) of the GLS estimators. The average estimation error of SGN at the final step ($\bar \phi_T=20.45$) is close to the estimation error of GLS ($19.31$).


%

\begin{table}[htbp]
	\centering
	\scalebox{0.8}{\begin{tabular}{r|rrll}
		\toprule 
		Regressor & \multicolumn{1}{l}{GLS} & \multicolumn{1}{l}{SGN} &   \quad CI(GLS) & \\
		\midrule
		Intercept & 0.5614 & 0.2204 & [0.4267, 0.6962] &\\
		Start dist & 0.372 & 0.1958 & [0.3132, 0.4308] &\\
		Diet(gi) & -0.474 &-0.3397 & [-0.6128, -0.3352]   &*  \\
		Diet(g) & -0.2781 & -0.0670 & [-0.4542, -0.1021]    &   \\
		Diet(i) & -0.4715 & -0.2993 & [-0.6448, -0.2981]  &*\\
		Latitude & -0.0946 &  -0.0548 & [-0.1425, -0.0467] &*\\
		Flock & -0.4886 & -0.3377 & [-0.5796, -0.3976] &\\
		Habitat & 0.0463 &  0.0280 & [0.0009, 0.0917] &*\\
		\bottomrule
		\multicolumn{5}{l}{
			\begin{minipage}{9cm}%
				 Table 1.  Results of GLS and SGN estimations, accounting for variation in FID in relation to starting distance, diet, latitude, flock size, habitat in European gregarious bird species. CI(GLS) is the $97\%$ confidence interval  of the GLS estimations and the SGN estimations are marked by $*$ if they fall into CI(GLS).
				
			\end{minipage}%
		}
		
		\label{tt}
		
	\end{tabular} }%
	\label{tab:addlabel}%
\end{table}%

\begin{figure}[htp]
	\centering
	\subfloat[]{\label{figur:1}\includegraphics[width=50mm]{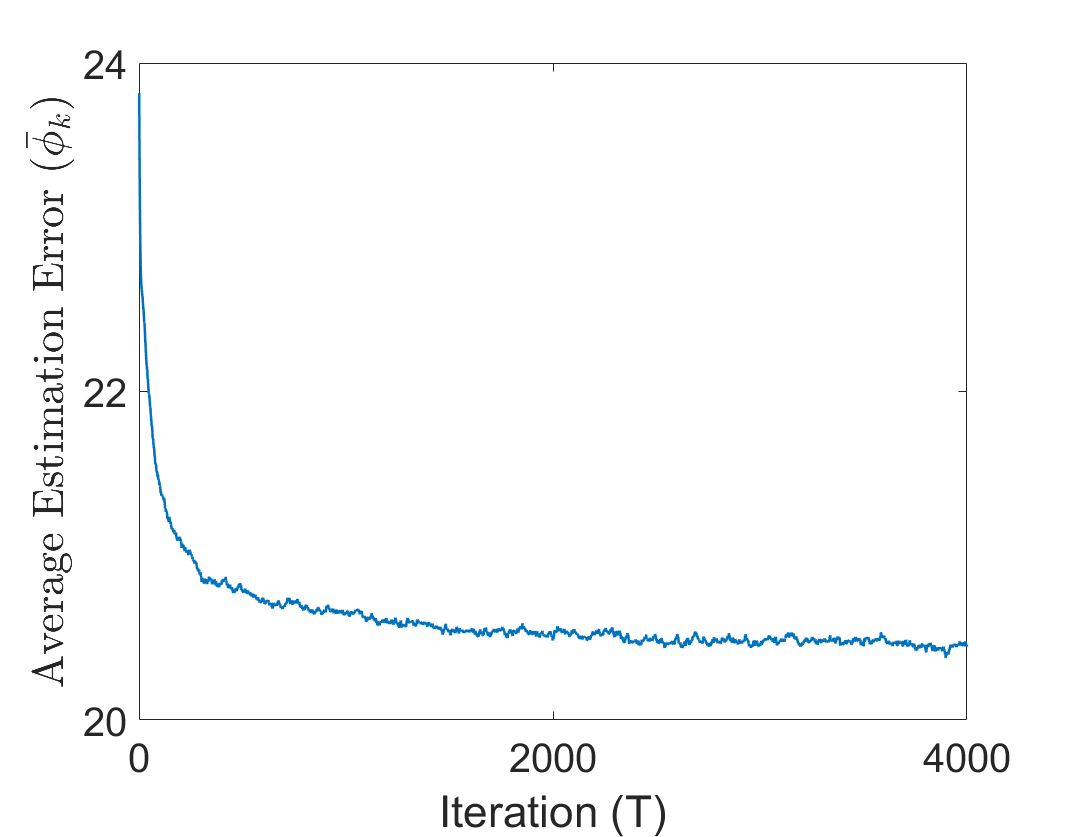}}\\
	
	\begin{minipage}{7cm}%
		\small Figure 3.  SGN average estimation error $\bar{\phi}_k$ at each iterations. 
	\end{minipage}%
	
\end{figure}

\subsection{Effects of Network Size and Connectivity}\label{ex1}
We consider a synthetically generated dataset that satisfies the assumptions of the setup in Section \ref{secsetup} in order to test the effects of (a) growing network size $N$ and (b) network connectivity. 

The numerical setup that is common to all experiments in this section is as follows. Each node receives 10 data points ($m=10$), and there are 5 features ($d=5$), i.e., $\X_i\in \R^{10\times5}$. We generate feature values $\X_i$ randomly using a normal distribution with mean $1$ and variance $0.1^2$. The individual noise term for each node comes from a zero-mean normal distribution with variance $\s_i^2$ randomly chosen between $0.001^2$-$0.1^2$. The matrix $\L_i$ is randomly created with its norm controlled between $0.003$-$0.3$ for all $i\in \mathcal{V}$. The output  $y_i$'s are generated according to \rf{y1} with $\w^*$ set as a vector of random integers. For all the experiments, we set  $\mu_i=1$, $\hat{\alpha}_i=\frac{1}{\trc{\O_i}}$ for each $i$, SGN stepsize as $\g=10^{-8}$, and the total number of iterations as $T=1000$. 

In the first experiment, we look at the effect of network size. We increase the data size by $m$ with the addition of each new node, i.e., $p = N m$, while keeping $\l=100$ fixed. Given an $N$ value, we run SGN for 10 trials. For each trial, we generate a random network with $N$ nodes by randomly keeping $\Upsilon_{N}=0.6$ fraction of the edges from the complete $N$-node network.

We measure the performance given $N$-node network using scaled final average estimation error: 
\[\Phi_s(N):=
\frac{1}{M}\sum_{l=1}^M\bar\phi_{T}^{(l)}(N),
\]
where $\bar\phi_{T}^{(l)}(N)$ is the average estimation error (as defined in  \eqref{phib})from the $l$th trial with $N$ nodes, and the corresponding estimation error is defined as in \eqref{phi}.
 We compute $\Phi_s(N)$ by averaging final estimation errors with $M=5$ and then dividing it by the scaling term $\bar \phi_T(N=5)$ and $\l = 0$, 
which makes sure $\Phi_s(N) \in (0,1)$ for all $N\in [5,150]$. Figure 4(a) shows a decrease of 15\% in average estimation error when we increase $N$ from 5 to 150 while keeping regularization penalty and connectivity the same.   

In the second experiment, we fix   $N=100$ and $\l=100$ while ranging $\Upsilon_{100}$ from $0$ to $1$. We recall $\Upsilon_{100}$ determines the fraction of edges kept from the complete $100$-node network. In Figure 4(b), the average estimation error is scaled with $\bar{\phi}_T(N=100)$ with $\l=0$ and  computed over 10 trials for a given $\Upsilon_{100}$.
Note that the case when $\Upsilon_{100}=0$, i.e., when the network is fully disconnected, is equivalent to having $\l=0$.
Here we observe that  addition of new edges improves the performance of SGN when $N$ is fixed. Indeed, the expected decrease in estimation error is about 20\% when we compare the fully disconnected case with the complete network.


Overall, the numerical experiments in this section support the consistency bound in Theorem 2 and the convergence to optimality in Lemma 2 for the SGN method. 

\begin{figure}[htp]
	\centering
	\subfloat[]{\label{figur:1}\includegraphics[width=50mm]{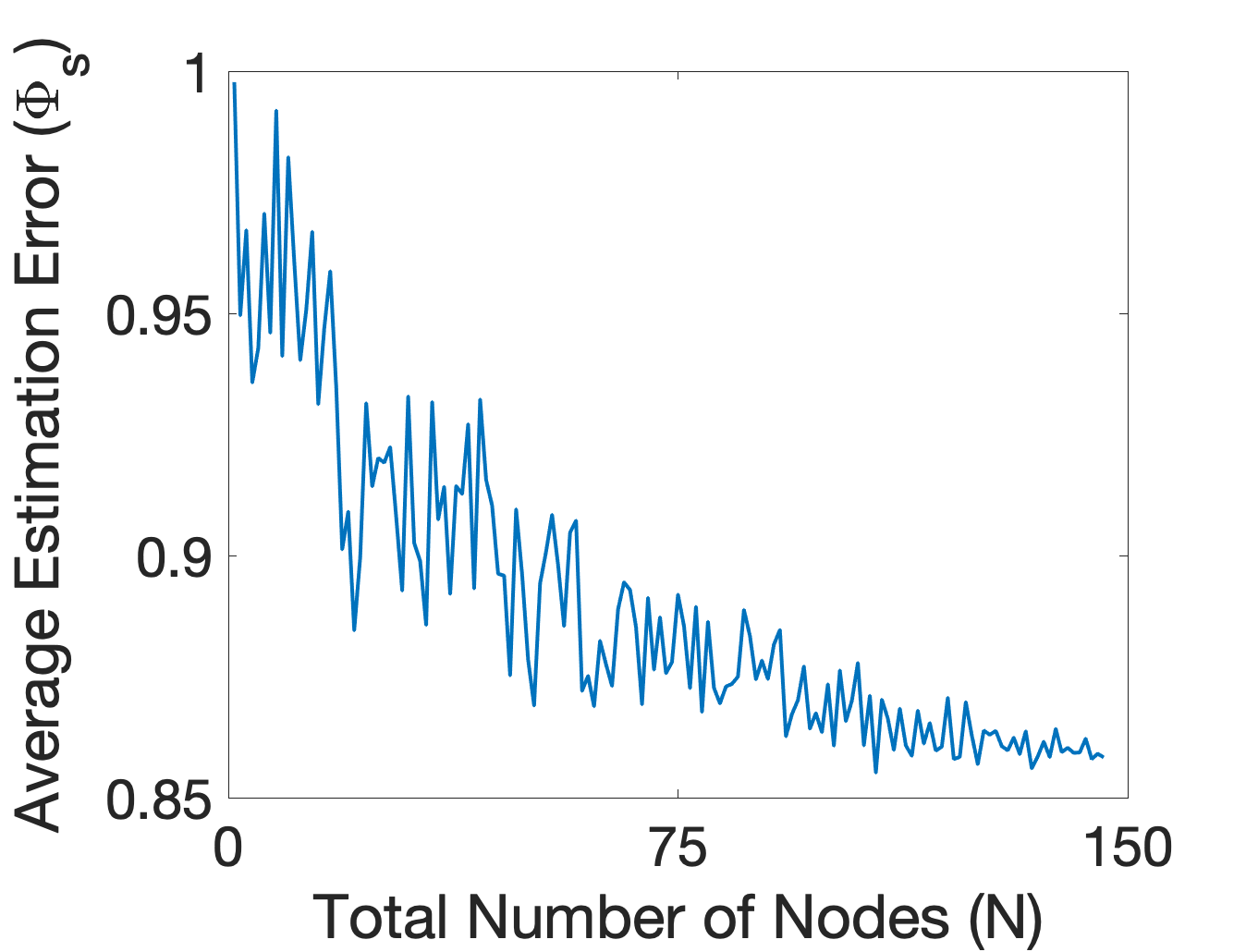}}
	\subfloat[]{\label{figur:3}\includegraphics[width=50mm]{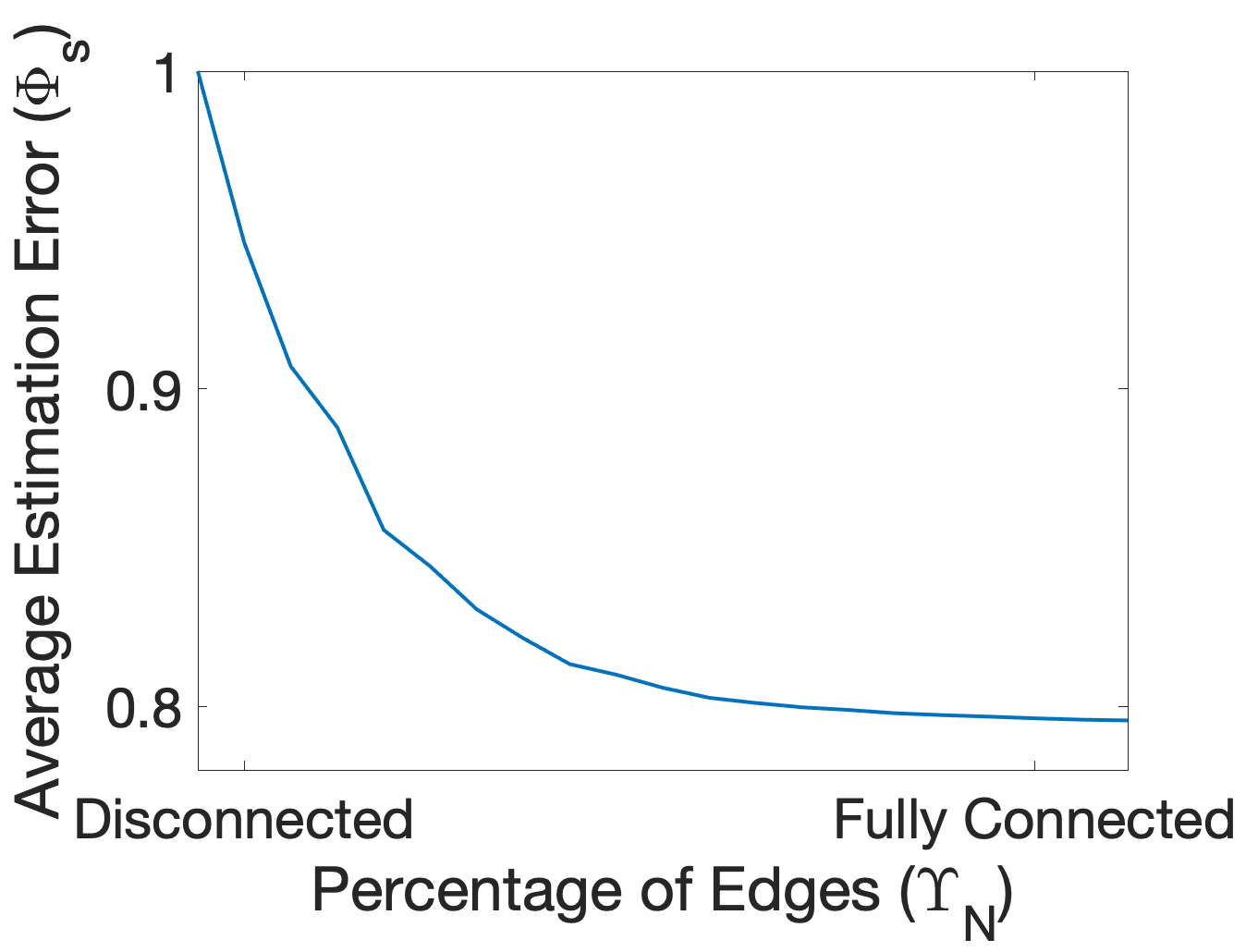}}
	\\
	\begin{minipage}{10cm}%
		\small Figure 4. Node size, connectivity and regularization parameter effects. (a) Average estimation error $\Phi_s$ as $N$ and $p$ increases. (b) Average estimation error $\Phi_s$ as the fraction of edges $\Upsilon_{N}$ increases. The network is disconnected when $\Upsilon_N=0$, and is fully connected when $\Upsilon_N=1$. 
	\end{minipage}
\end{figure}


\section{Conclusions}

The ever-increasing dimension of data and the size of datasets have introduced new challenges to centralized estimation. For example, limited bandwidth in the current networking infrastructure may not satisfy the demands for transmitting high-volume datasets to a central location.
Hence, it is of interest to study alternatives to centralized estimation. 

In this paper we consider a distributed architecture for learning a linear model via generalized least squares by relying on a network of interconnected ``local" learners. In the proposed distributed scheme, each computer (or local learner) is assigned a dataset and {\em asynchronously} implements stochastic gradient updates based upon a sample. To ensure robust estimation, a network regularization term that penalizes models with high {\em local} variability is used. 
Unlike other model averaging schemes based upon a synchronized step, the proposed scheme implements local model averaging continuously and asynchronously. We provide finite-time performance guarantees on consistency. We illustrate the application of the proposed method for estimation in a Markov Random Field with synthetic datasets and a real dataset from ecology.


\newpage
\section*{Appendix}
\subsection{Notation}

\begin{table}[htbp]
	\centering
	\scalebox{0.8}{\begin{tabular}{l|l}
			\toprule 
			Symbol &Meaning\\
			\midrule
			N & number of computing nodes\\
			$m (d) $& row (column) space dimension of the data subsets\\
			$\w^*$ &  the ground truth  coefficient vector \\
			$\w_i$ & estimated coefficient vector  at node $i$\\
			$\hat{\w}_t$ & weighted average coefficient vector at time $t$\\
			$\ve_i$ & individual noise vector specific to dataset at node  $i$\\
			$\xi$ & common noise that affects all nodes\\
			$\alpha_{i}$& weight given to  node $i$\\ 
			$\L_i$ &  diagonal matrix represents the influence of $\xi$ on node $i$\\
			$\O_i$ & covariance matrix of the error term for $\y_i$\\
			$\hat{A}$ &  the weighted adjacent matrix \\
			$f_i(\w_i)$ & local measure of model fit\\
			$\rho_i(\w_i)$ &  network  regularization  penalty\\
			$\delta$ & penalty parameter \\
			$\g$ & stepsize\\
			$\tau_i, \v_i $&$\tau_i=\s_i\sqrt{ \g\mu_i}$ and $\v=\sqrt{ \g\mu_i}$\\
			$B\itt$, $B_t$& the  standard m-dimensional  Brownian  Motion\\& approximating the  individual noise associated with \\&node $i$ and the common noise\\ 
			$\lambda(A)$ & eigenvalue of matrix $A$,\\& $\lambda_2$ is the second smallest eigenvalue of $A$.\\
			$\eta $& Lipschitz parameter of $g_i$'s, $\eta=\max_i\norm{\X_i\trs\oi_i\X_i}$ \\
			$	\k $ & $\k=\min_i \k_i$, where $\k_i$ is the strong convexity parameter \\&of $f_i$\\
			$	\mu_i $&stream parameter for gradient estimate at node $i$,\\ & $\mu=\min_i \mu_i$, and $\mu'=\max_i \mu_i$\\
			$	\beta $&stream parameter for communication gradient estimate\\
			$h_t$ &$\min_{i\in \mathcal{V}}g_k(\hz)$\\
			$q_t$ &$\max_{k\in \mathcal{V}}g_k(\hz)$\\
			$A_{i,k}$ &$\mathbf{1}\trs( \X_i\trs\oi_i\L_i \circ \X_k\trs\oi_k \L_k) \mathbf{1}$ \\
			\bottomrule
	\end{tabular} }%
	\label{tab:addlabel}%
\end{table}%
\subsection{Technical preliminaries}\label{sec_tech}
We will make use of the following definitions and Ito's lemma.
Let $f: \mathcal{X} \to \mathcal{S}$ be a function with gradient $\tr f(x)$.
\begin{definition} \label{strong}
	$f$ is called a Lipschitz function if there exists a constant $\eta>0$, such that 
	$$\norm{ f(x_1)-f(x_2)}\leq \eta \norm{x_1-x_2}$$
	for some all $x_1$, $x_2\in \mathcal{X}$.
\end{definition}

\begin{definition}\label{strict}
	Twice differentiable function $f$ is said to be $\k-$strongly convex, if
	$$(\tr f(x_1)-\tr f(x_2))^T(x_1-x_2)\geq \frac{\k}{2}\norm{x_1-x_2}^2$$
	for some $\k>0$ and all $x_1$, $x_2\in \mathcal{X}$. Or equivalently, $\tr^2f(x) \succeq\k I$  for all $x\in \mathcal{X}$, i.e., $a_{min}(\tr^2f(x))\geq \k$.
	where $\tr^2f(x) $ is the Hessian matrix, and $a_{min}(\cdot)$ is the minimum eigenvalue.
\end{definition}

\begin{lemma}{Multidimensional Ito Lemma}\label{ito}\cite{oksendal2003stochastic}\\
	Let \[dX_t=\mathbf{u}dt+VdB_t\]
	be an m-dimensional Ito process, where $\mathbf{u}$ is a vector of length $m$,  $V$ is an $m\times m$ matrix, and $B_t$ is a $m$-dimensional  Brownian motions. Let $g_{t,x}$ be a twice differentiable map from $\R^m$ into $\R$. Then the process
	\[Y_t=g_{t,X_t}\]
	is again an  Ito process with 
	\[dY_t=\frac{\partial g}{\partial t}dt+\sum_{i}\frac{\partial g}{\partial x_i}dX_i +\frac{1}{2}\sum_{i,j}\frac{\partial^2g}{\partial x_i \partial x_j}dX_i dX_j,\]
	where $dX_idX_j$ is computed using rules $dtdt=0$, $dtdB_i=0$, $(dB_i)^2=mdt$, $B\itt B_t=0$,  and $B_iB_j=0$ for all $i\neq j$.
\end{lemma}

\subsection{Continuous Time Representation of the Approximation Dynamics }\label{A1}	
In this section, we will derive the formula of $d\w\itt$ by rewriting the scheme \rf{pw} in the form of the summation of previous steps, and approximate the noise terms by standard $m$-dimensional Brownian motions and the rest by integrals. Then $d\w\itt$ can be approximated by the differential form of a stochastic Ito integral.  We initially assume both noise terms have zero-mean Gaussian distribution: $\xi_k\sim \mathcal{N}_m(\mathbf{0},  \mathbf{I}_m)$ and $\ve_{i,k}\sim \mathcal{N}_m(\mathbf{0}, \s_i^2 \mathbf{I}_m)$ for all $i$.  Later we show via Central Limit Theorem that this approximation also holds for general distributions.

\subsubsection{Approximation with Gaussian distributed noise term}\label{gau}

For each node $i\in\mathcal{V}$, we rewrite the scheme \rf{pw} as: 
\eqs{02}{\w_{i,t}=&\w_{i,0}-
	\g\sum_{k=1}^{N_{g,i}(t/\g)}g_{i,k}-
	 \g\l\sum_{k=1}^{N_{r}(t/\g)} \tr \rho_{i,k}\\&
	 + \g\X_{i}\trs\oi_i\sum_{k=1}^{N_{g,i}(t/\g)}\ve_{i,k} + \g\X_{i}\trs \oi_i\Lambda_i\sum_{k=1}^{N_{g,i}(t/\g)}\xi_k.  }
Consider the second and the third term in \rf{02}. If $\g\ll \mu_i$, it follows that:
\eqs{03}{ \tg \sum_{k=1}^{N_{g,i}(t/\g)} g_{i,k}=&\frac{\g N_{g,i}(t/\g)}{t}\sum_{k=1}^{N_{g,i}(t/\g)} g_{i,k}\frac{t}{N_{g,i}(t/\g)}\\
\approx&\frac{\g N_{g,i}(t/\g)}{t}\sum_{k=1}^{N_{g,i}(t/\g)} g_{i,k}\frac{\g}{\mu_i}\\
\approx &\mu_i \int_{0}^{t} g_{i,s} ds.}  
Similarly,
\eqs{04}{ \g \sum_{k=1}^{N_{r}(t/\g)}\tr\rho_{i,k}=&\frac{\g N_{r}(t/\g)}{t}\sum_{k=1}^{N_{r}(t/\g)} \tr\rho_{i,k}\frac{t}{N_{r}(t/\g)}\\
	 \approx&\beta \int_{0}^{t}\tr\rho_{i,s} ds.}

Now consider the individual noise. We assume $\ve_{i,k}\sim \mathcal{N}_m(\mathbf{0}, \s_i^2 \mathbf{I})$,  all components of $\ve_{i,k}$ are independent and it is enough to illustrate one-dimension approximation. Let $\ve^{(q)}_{i,k}$ be the $q$th dimension of $\ve_{i,k}$, and for all $q\in \mathcal{D}=\{ 1,\dots, m\}$, it follows that
\[\E\big[ \sum_{k=1}^{N_{g,i}(t/\g)}\g\ve^{(q)}_{i,k}\big]= 0, \]
\[  \var\big[ \sum_{k=1}^{N_{g,i}(t/\g)} \g\ve^{(q)}_{i,l}\big]=\frac{\g N_{g,i}(t/\g)}{t} \g\s_i^2 t\approx\g\mu_i\s_i^2 t. \]
We approximate $ \sum_{k=1}^{N_{g,i}(t/\g)}  \ve_{i,k}$ by a standard $m$-dimensional Brownian Motion $B\itt$:
\al{ \sum_{k=1}^{N_{g,i}(t/\g)} \ve_{i,l}\approx \s_i \sqrt{ \g\mu_i} B_{i,t} =\tau_i   B_{i,t},         }
where $\tau_i=\s_i\sqrt{ \g\mu_i}$. Hence the individual noise term in \rf{02} can be approximated as 
\eq{ind}{	\g \X_{i}\trs\oi_i\sum_{k=1}^{N_{g,i}(t/\g)}\ve_{i,l}\approx \tau_i \X_i\trs \oi_i B\itt.}
Similar to the proof of the individual noise approximation, let  $\xi_{i,k}^{(q)}$ be the $q$th dimension of $\xi_{i,k}$, for $q \in \mathcal{D}$, the it follows that
\[ \E\big[ \sum_{k=1}^{N_{g,i}(t/\g)}\g\xi_{i,l}^{(q)}  \Big]=  0,\]
\[  \var\big[\sum_{ l<t } \g\xi_{i,l}^{(q)} \Big]=  \frac{\g N_{g,i}(t/\g) }{ t}\g t\approx \mu_i\g t.\]\
Let $\v_i=\sqrt{ \g\mu_i}$, then we approximate the common noise term in \rf{02} by a standard $m$-dimensional Brownian Motion $B_t$:
\eq{06}{ \g\X_{i}\trs\oi_i\L_i\sum_{k=1}^{N_{g,i}(t/\g)}\xi_k	 \approx \v_i \X_i\oi_i\L_i B_t.   }

Substituting \rf{03},  \rf{04} and \rf{06} to the corresponding terms in \rf{02}, $\w\itt$ approximately satisfies the following stochastic Ito integral:
\al{  \w\itt=&\w_{i,0} -\mu_i \int_{0}^{t} g_{i,s} ds
	- \l \b  \int_{0}^{t}\tr \rho_{i,s} ds	\\&+\tau_i \int_{0}^{t} \X_i\trs \oi_i d B_{i,s}  + \v_i\int_{0}^{t}  \X_i\trs \oi_i\L_i d B_s .     }
Taking the derivative of the above equation, we get \rf{dpw}.


\subsubsection{Approximation with noise from general distributions}\label{gener}
Let the individual noise $\ve_i$'s follow a distribution of  expected value $\mathbf{0}$ and covariance matrix $\s_i^2 \mathbf{I}$, and the noise vectors are i.i.d within the same subset.
The common noise vectors are i.i.d, and follow a distribution with zero mean and identity covariance.  

We can see $\sum_{k=1}^{N_{g,i}(t/\g)}  \xi_{k}/N_{g,i}(t/\g)$ and $\sum_{k=1}^{N_{g,i}(t/\g)}  \ve_{i,k}/N_{g,i}(t/\g)$ as the average of  sequences of i.i.d. random variables. For all $i\in\mathcal{V}$ and   $q\in \mathcal{D}$, by the Central Limit Theorem, 
\al{&\frac{ \sum_{k=1}^{N_{g,i}(t/\g)}  \ve^{(q)}_{i,k}}{\sqrt{N_{g,i}(t/\g)}}\sim \mathcal{N}(0,\s_i^2), \quad
\frac{\sum_{k=1}^{N_{g,i}(t/\g)}  \xi_{k}^{(q)}}{\sqrt{N_{g,i}(t/\g)}}\sim \mathcal{N}(0, 1).   }
It also follows that
\[\var\Big[\sqrt{N_{g,i}(t/\g)}\frac{\sum_{k=1}^{N_{g,i}(t/\g)}  \xi_{k}^{(q)}}{\sqrt{N_{g,i}(t/\g)}} \Big]=\g\mu_i t , \]
\[\var\Big[ \sqrt{N_{g,i}(t/\g)}\frac{ \sum_{k=1}^{N_{g,i}(t/\g)}  \ve^{(q)}_{i,k}}{\sqrt{N_{g,i}(t/\g)}}\Big]= \g\mu_i\s_i^2 t, \]
Thus we would have the same noise Brownian approximation as in \rf{ind} and \rf{06}, and the rest of steps follow as in \rf{gau}. Hence we showed the continuous representation of $\w_{i,t}$ dynamics with general noise distribution.

\subsection{Proof of Theorem 1}\label{C1}
\subsubsection{The differential form of the regularity measure  }\label{B1}
The following lemma provides the differential form of the regularity measure. We apply the Ito's lemma (Lemma 3) to $d V\itt$ and then take the weighted average of $d V\itt$'s to obtain $d\bar{V}_t$.

\begin{lemma}
The regularity measure $\bar{V}_t $ satisfies
\eqs{pV}{d\bar{V}_t\leq&-\sumbs \mu_ig\itt\trs \ee\itt dt -  \sumbss{\l\b}  \tr\rho\itt\trs \ee\itt  dt \\&+K_1d\td{B}_t +C_1dt	}
where $K_1d\td{B}_t$ is the summation of Ito terms,
\al{K_1 d\td{B}_t=\sum_{i=1}^N\a_i K_{1,i}d\td{B}_t,}   	with $K_{1,i}d\td{B}_t$  for $i\in\mathcal{V}$ defined as,
\al{K_{1,i}d\td{B}_t=&\tau_i \X_ i\trs\oi_idB\itt\trs\ee\itt+ \v_i\X_i\trs\oi_i \Lambda_i  dB_t \trs \ee\itt.   }
\end{lemma}
\proof By definition of $\hz$, we take the weighted average of \rf{dpw} and obtain
\eqs{dpww}{d\hz=&-\sumbks \mu_kg\kt dt   -\sumbkss{\l\b}  \tr \rho\kt dt\\& +
\sum_{k=1}^N\a_k\tau_k\X_k\trs \oi_k dB\kt  +\sumbks\v_k\X_k\trs\oi_k\L_k dB_t.     }


We subtract \rf{dpww} from \rf{dpw}, and by definition of $\ee\itt$, it follows that
\eqs{22}{d\ee\itt=&d\w\itt-d\hz\\
=&  \Big(-\mu_ig\itt+  \sumbks{\mu_k} g \kt\Big)  dt    \\                                           
&+ \l\b\Big(-  \tr\rho\itt +\sumbk  \tr\rho\kt\Big)   dt \\
&+ \Big( \tau_i \X_i\trs\oi_idB\itt-\sumbks{\tau_k} \X_k\trs\oi_k dB\kt    \Big)			\\
&+\Big( \v_i\X_{i}\trs\oi_i\Lambda_i - \sumbks{\v_k}\X_{k}\trs\oi_k\L_k \Big)dB_t.
}
Note that for  $d$-by-$m$ matrices $C=[c_1,\dots,c_d]$ and $Q=[q_1,\dots,q_d]$, $CdB_t\cdot CdB_t=\norm{C}_F^2dt$, and  $CdB_t\cdot  QdB_t=\mathbf{1}\trs (C\circ Q) \mathbf{1}$, where  ``$\mathbf{1}$" is a vector of all ones and  ``$\circ$" denotes the Hadamard product.
Then by \rf{22}, the inner product of two $d\ee\itt$ is 
\eqs{23}{d\ee\itt\cdot d\ee\itt=  C_{1,i} dt, }
and hence $C_{1,i}$ is positive.\\
Applying Ito's lemma to $dV\itt$, we obtain
\eqs{vvp}{dV\itt=&\ee\itt \cdot d\ee\itt+\frac{1}{2}d\ee\itt\cdot d\ee\itt\\	
\leq&  \Big(-\mu_ig\itt+  \mu'\sumbks g \kt\Big)\trs \ee\itt     dt  \\&
+  \l\b\Big(-  \tr\rho\itt +\sumbk  \tr\rho\kt\Big) \trs\ee\itt dt \\
&+\frac{1}{2}C_{i,1} dt+ K_{1,i}d\td{B}_t- K_{1}\td{B}_t,
}
where $\mu'=\max_i \mu_i$. Note that the quadratic variation of $\ee\itt$ is as follows, 
\[ \vb{\ee\itt}_{t}=\E\int_0^{t} \frac{1}{2}C_{1,i} ds\to \var(\ee\itt).\]
Hence, $\frac{t}{2}C\itt$ describes the variation of $\ee\itt$, and  $C_1$ explains the variation of the weighted approximation difference (to the ensemble solution).

To obtain the differential form of the regularity measure, we take the weighted average of \rf{vvp}. Because of \rf{s0}, the terms with double summation (weighting)  vanish, and \eqref{pV} follows. 
\endproof

\subsubsection{Proof of Theorem 1}
In the following proof, we first obtain an upper bound of $d\bar{V}_t$ given in \rf{pV} using the $\eta$-Lipschitz continuity of the gradient $g_i$ and the properties of the Laplacian matrix. Second, we integrate and take the expectation of the obtained bound to get the desired result.

Consider the first term of \rf{pV}, let $h_t=\min_{i\in \mathcal{V}}g_{i,t}(\hz)$. By \rf{s0}, we can add a zero-valued term $h_t\trs\sum_{i=1}^{N}\a_i \ee\itt$ to the equation, and by the strong convexity of $g_i$ \rf{conv}, we can obtain the following inequality,
\eqs{36}{ -\sum_{i=1}^N \alpha_i g\itt\trs \ee\itt &\leq -\sumbss{\mu} (g\itt-h_t)\trs \ee\itt\\&\leq-	\sumbss{\mu} (g\itt-g_{i,t}(\hz))\trs \ee\itt \\ &\leq -\mu\sumbs{\k_i\norm{\ee\itt}^2}  \leq -2\k \mu\bar{V}_t,}
where $\mu=\min_i \mu_i$ and $\k=\min_i\k_i$. 
The first inequality above follows from the definition of $h_t$, and  the second inequality is by the strict convexity of the gradient $g_i$. 

Now we consider the second term of \rf{pV}. Define the vector $\ee_t=[\ee_{1,t}^T, \dots,\ee_{N,t}^T ]^T$ and the matrix $\hat{\*L}= \hat{L}\otimes I_m$, where $\otimes$ is the Kronecker product. Using these definitions, we have the following obervation:

\eqs{31}{-\sum_{i=1}^{N}\sum_{j\neq i}& \hat{\alpha}_i\hat{\alpha}_j a_{i,j}(\w\itt-\w\jt)\trs \ee\itt \\
=&-\sum_{i=1}^{N}\sum^N _{j= i} \hat{\alpha}_i\hat{\alpha}_j a_{i,j}(\ee\itt  -\ee\jt)\trs\ee\itt
\\=&
\sum_{i=1}^{N}\sum^N _{j= 1} \hat{l}_{i,j}(\ee\itt  -\ee\jt)\trs\ee\itt
\\=&\sum_{i=1}^{N}\norm{\ee\itt}^2\sum^N _{j= 1} \hat{l}_{i,j}
-\sum_{i=1}^{N}\sum^N _{j= i} \hat{l}_{i,j}\ee\jt\trs\ee\itt\\
=&-\ee_t\trs \hat{\*L} \ee_t, }
where $\hat{l}_{i,j}$ is the $(i,j)$th entry of $\hat{L}$. The first equality is by adding and subtracting $\hz$, and adding the zero-valued term $\hat{\a}_i\hat{\a}_ja_{i,i}(\ee\itt  -\ee\itt)\trs\ee\itt$ (the $(i,i)$th term). The second equality follows from
 $\hat l_{ij}=-\hat{A}_{ij}$ for all $i\neq j$, and the $(i,i)$th term is zero-valued. The final equality follows that  the column sums of $\hat{L}$ is zero.
The second smallest eigenvalue $\hl>0$ satisfies (see \cite{godsil2001g}),
\[
\min_{x\neq0,\, \boldmath{1}\trs x=0}\frac{(x\trs \hat{L} x)}{\norm{x}^2}=\hl.
\]
 Thus, we have
\eq{33}  {-\ee_t^T \hat{\*L} \ee_t\leq-\hl  \sum_{i=1}^{N}\norm{\ee\itt}^2.}
We assume that $\hat{\a}_i=c\a_i$, it follows that,
 \eqs{38}{
-\l\beta \sum_{i=1}^{N} \alpha_i\tr\rho\itt\trs\ee\itt &=
-\frac{\l\b}{c}\sum_{i=1}^N \sum_{j\neq i} \hat{\a}_i\hat{\a}_j a_{i,j}(\w\itt-\w\jt)\trs \ee\itt\\
 &\leq -\frac{\l\b}{c} \hl \sum_{i=1}^{N} \norm{\ee\itt}^2 \\
	&\leq  -\frac{2\l\b}{c} \hl  \sum_{i=1}^{N} \a_i V\itt \\ 
	& =-2\l \hat{\beta}\hl  \bar{V}_t.}
where $\hat{\beta}=\beta/c$.
The second inequality follows by combining \rf{31} and \rf{33}, the third inequality follows by scaling the terms with $\a_i<1$,  and the final equality is by the definition of $\bar V_t$.
%
An upper bound for $d\bar{V_t}$ follows from $ \rf{36}$ and $\rf{38}$,
\eqs{310}{d\bar{V}_t
	&\leq -2(\mu\k+\l\hb\hl)   \bar{V}_t  dt+ K_1d\td{B}_t +C_1dt	\\}
Now consider the derivative of $e^{2(\mu\k+\l\b\hl) t}\bar{V}_t$,
\eqs{311}{ d&(e^{2(\mu\k+\l\hb\hl) t}\bar{V}_t )\\ = &e^{2(\mu\k+\l\hb\hl) t}d\bar{V}_t +2(\mu\k+\l\hb\hl) e^{2(\mu\k+\l\hb\hl) t}\bar{V}_tdt \\
	\leq& e^{2(\mu\k+\l\hb\hl) t}C_1dt +e^{2(\mu\k+\l\hb\hl) t} K_1d\td{B}_t,}
where the inequality follows by using \rf{310} for each $d \bar V_t$ term. Integrating both sides of the inequality in \rf{311},
\eqs{3121}{\bar{V}_t\leq &e^{-2(\mu\k+\l\hb\hl) t}\bar{V}_0+ \frac{ C_1}{2(\mu\k+\l\hb\hl)}( 1-e^{-2(\mu\k+\l\hb\hl) t})\\&+e^{-2(\mu\k+\l\hb\hl) t}\int_{0}^{t}e^{2(\mu\k+\l\hb\hl) s}K_1d\td{B}_s.     }
Since the stochastic integral is a martingale,
\[ 
\E\Big[\int_{0}^{t}e^{2(\mu\k+\l\hb\hl) s}K_1d\td{B}_s\Big]=0.
\]
We obtain the desired upper bound by taking the expectation on both sides of \rf{3121}.
In the long run, as  $t\to \infty$, the exponential terms will vanish, and the upper bound of the regularity measure follows.

\subsection{Proof of Theorem 2}\label{D1}
The proof follows a similar outline as Theorem 1. We start with Ito's Lemma to get the stochastic dynamics form of $d U_t$ and then introduce an auxiliary variable $W_ t$ that depends on both $U_t$ and $\bar V_t$ to bound $E[U_t]$.

We apply Ito's lemma to $dU_t$, and use the identity in \rf{s0} and the differential form of $\hat{\w}_t$ in \rf{dpww} to get the following form,
\eqs{411}{dU_t=&(\hz-\tw)\cdot d(\hz-\tw)+\frac{1}{2}d(\hat{\w}_{t}-\tw)\cdot d(\hz-\tw)\\
	=&-\sumbs{\mu_k} g\kt\trs(\hz-\tw)dt+\\& \sum_{k=1}^N\a_k\tau_k\Big( \X_k\trs \oi_k dB\kt   \Big)\trs(\hz-\tw)       \\&
	+ \sumbks\v_k(\X_k\trs\oi_k\L_k dB_t)\trs(\hz-\tw)  +   \\& \frac{1}{2}\Big(\sum_{k=1}^{N}\a_k^2\tau_k^2\norm{\X_k\trs\oi_k }_F^2+\sum_{k=1}^N\sum_{j=1}^{N}\a_k\a_j\v_k\v_jA_{k,j}   \Big)dt\\}
We separate the first term of \rf{411} using the identity $\hat{\w}_{t}-\tw = (\w\kt-\tw )-  \ee_{k,t}$, and rearrange constant and Ito terms to get
\eqs{41}{dU_t	=&-\sumbs{\mu_k} g\kt\trs(\w\kt-\tw)dt + \\&\sumbs{\mu_k} g\kt\trs  e_{kt}dt + K_2d\td{B}_t+C_2dt 
}
where the summation term $C_2$ is defined in \eqref{c2} and $\ub$ is given as, 
\al{K_2d\td{B}_t= &\sum_{k=1}^N\a_k\tau_k(\hz-\tw) \trs \X_k\trs\oi_k  dB\kt  \\&   +  \sumbks\v_k(\hz-\tw) \trs\X_k\trs\oi_k\L_k dB_t. }
Note that the quadratic variation of $\hz-\w^*$ is as follows, 
\[\vb{\hz-\w^*}_{t}=\E\int_0^{t} C_2 ds\to \var(\hz-\w^*).\]
Hence $tC_2$ describes the variation of $\hz-\w^*$.

We have the following upper bound on the first term of \eqref{41},
\eqs{42}{-\sumbs{\mu_k}& g\kt\trs(\w\kt-\tw)\\\leq& -\sumbss{\mu}( g\kt-g(\tw))\trs(\w\kt-\tw)\\
	\leq& -2\k\mu \sum_{k=1}^{N}\frac{\a_i}{2}\norm{\w\kt-\hz+\hz-\tw}^2
	\\=&-2\k\mu (\bar{V}_t+U_t).
}
The first equality is obtained by subtracting the zero-valued term $g(\tw)$ from $g\kt$. The second inequality is by strong convexity of the gradients ($g\kt$), and the last equality follows from \rf{d}.

Now consider the second term of \rf{41}. Let $q_t=\max_{k\in \mathcal{V}}g_k(\hz)$, then it follows that
\eqs{43}{ & \sumbs{\mu_k} g\kt\trs e\kt\leq\sumbss{\mu'} (g\kt-q_t)\trs e\kt \\ &\leq2\mu' \sum_{k=1}^{N}\frac{\a_i}{2} (  g\kt-g_k(\hz)     )\trs e\kt \leq 2\mu'\eta \bar{V}_t. }
The equality is obtained by subtracting the zero-valued term $q_t \sum_{k=1}^{N}\a_k\ee\kt$. The first inequality follows by the definition of $q_t$, and the second inequality is by the Lipschitz continuity of $g\kt$'s. 
By   \rf{41}$-$\rf{43}, we can obtain an upper bound of $dU_t$,
\eqs{du}{dU_t\leq &-2\mu\k  U_tdt +2(\eta\mu'-\k\mu) \bar{V}_tdt\\&+C_2dt+   K_2d\td{B}_t.  }

We construct an auxiliary variable $W_t$ and then follow similar steps as in the proof of Theorem 1:  integrate the upper bound of $dW_t$ and then take expectation. Now, 
define 
\[ W_t=U_t+\frac{\eta\mu'-\k\mu}{\l\hb\hl  }\bar{V}_t. \]
The differential of $W_t$ can be obtained as follows,
\eqs{dw}{dW_t\leq &-2\mu\k  U_tdt +\Big(2(\eta\mu'-\k\mu) \bar{V}_t+C_2    \Big)dt\\&+K_2d\td{B}_t +\frac{\eta\mu'-\k\mu}{\l\hb\hl   }\Big(-2(\k\mu +\l\hb\hl)\bar{V}_tdt\\&+ K_1d\td{B}_t+C_1dt  \Big)\\
	=& -2\mu \k W_tdt+\Big(\frac{\eta\mu'-\k\mu}{\l\hb\hl   }C_1+C_2\Big)dt + \\&\big(\frac{\eta\mu'-\k\mu}{\l\hb\hl   }K_1+K_2 \big)d\td{B}_t.}
The inequality is obtained by plugging in the bounds for $d \bar{V}_t$ in \rf{310} and for $dU_t$ in \rf{du} into $dW_t$. The first equality follows by rearranging terms, and the second one is obtained by combining terms that are equivalent to $W_t$. 

Consider the derivative of $e^{2\mu\k t}W_t$ and plug in the upper bound of $dW_t$ in \rf{dw} to obtain, 
\eqs{46}{ d(e^{2\mu\k t}W_t) =& e^{2\mu\k t}dW_t+ 2\mu\k e^{2\mu\k t}W_t  dt  \\
	\leq& e^{2\mu\k t}\Big(\frac{\eta\mu'-\k\mu}{\l\hb\hl   }C_1+C_2 \Big)dt +\\& e^{2\mu\k t}\big(\frac{\eta\mu'-\k\mu}{\l\hb\hl   }K_1+K_2 \big)d\td{B}_t
}
The following inequality is obtained by integrating both sides of \rf{46}, 
\eqs{47}{W_t\leq&  e^{-2\mu\k t}\Big[W_0    + \int_{0}^{t}e^{2\mu\k s}\Big(\frac{\eta\mu'-\k\mu}{\l\hb\hl   }C_1+C_2 \Big)ds \\&+  \int_{0}^{t} e^{2\mu\k s}\big(\frac{(\eta\mu'-\k\mu)}{\l\hb\hl   }K_1+K_2 \big)d\td{B}_s\Big]\\
	= &e^{-2\mu\k t}W_0    +\frac{1}{2\mu\k}\Big(\frac{\eta\mu'-\k\mu}{\l\hb\hl   }C_1+C_2 \Big)\Big(1-e^{-2\mu\k t}\Big)   \\&+ e^{-2\mu\k t}\int_{0}^{t} e^{2\mu\k s}\big(\frac{(\eta\mu'-\k\mu)}{\l\hb\hl   }K_1+K_2 \big)d\td{B}_s.
}
We assume that all nodes have the same initial estimate, i.e., $\w_{i,0}=\w_{j,0}$ for all $i,j$. Then, $\bar{V}_0=0$, which means $W_0=U_0$. Note that the stochastic integration of the Ito term is martingale and hence the expectation is zero. Taking expectation on both sides of \rf{47}, then
\eq{48}{ \E[W_t]\leq  e^{-2\mu\k t}U_0 + \frac{1}{2\mu\k}\Big(\frac{\eta\mu'-\k\mu}{\l\hb\hl  }C_1+\js \Big)\Big(1-e^{-2\mu\k t}\Big).       }
Since $\eta\mu'-\k\mu>0$,  we have 
$$\E[U_t]\leq \E[W_t].$$
Hence the right-hand side of \rf{48} is also the upper bound of $\E[U_t]$.
In the long run, as $t\to \infty$,  the exponential terms  vanish and the upper bound of the consistency measure follows.

\subsection{Proof of Lemma \ref{u} }\label{F1}

 We will find upper bounds for \rf{eu}, or more specifically, upper bounds for  the constants term  $C_1/\l \hb\hl$ (defined in  \eqref{c1}-\eqref{cts}) and $\js$ (defined in \eqref{c2}), and further prove that these bounds are functions of $\g$ and can be controlled as small as needed. With a slight abuse of notation, we refer to $N$ as the total number of nodes. 



In this lemma, we assume $c=\sum_{l=1}^N \hat{\a}_l$, i.e., $\a_i=\frac{\hat{\a}_i}{\sum_{l=1}^N\hat{\a}_l}$. We assume that $ 0<\e\leq \s_i^2$, and denote $\hat{\alpha} := \min_{i \in \mathcal{V}} \hat{\alpha}_{i}$, and $\hat{\alpha}' :=\max_{i \in \mathcal{V}} \hat{\alpha}_{i}$. Let $\o_1<\o_{i,k}<\o_2$ for all $i\in \mathcal{V}$ and $k\in{1,\dots,m}$.
Let $d$ be the scalar such that $d\hl=\lambda_2$, by choosing $\l \hl\sim N$ ( $\l d\lambda_2\sim N$, i.e., network connectivity is preserved) and $ \g \sim \e^3$, we prove that the difference between the ensemble average estimate and the ground truth is $O(\g^{1/2})$ as $N \rightarrow \infty$. 
We start by discussing the bound for $C_2$ and then use the results to construct the bound for $C_1/\l \hb\hl$.  For notational simplicity,  we substitute $\l\hb \hl$ by  $N$ in the following discussion.

\subsubsection{Limiting property of $ C_2$ }
Consider the first summation term of $\js$ inside the parentheses.  For $i\in \mathcal{V}$, rewrite  the data subsets and the matrices as  $X_i\trs=[\x_{i,1},\dots, \x_{i,m}]$ and $\L_i=\text{diag}(\o_{i,1},\dots, \o_{i,m})$, 
let $x_m$ be the largest element among the data matrix $X$, then 
 \al{\tau_i^2 \hat{\a}_i^2&\norm{\X_i\oi_i}^2_F=\s_i^2\hat{\a}_i^2\g\mu_i \norm{\X_i\oi_i}_F^2\\<& mx_m^2\sum_{k=1}^{m}\frac{\s_i^2\hat{\a}_i^2\g\mu_i}{(\s_i^2+\o_{i,k}^2)^2}<\frac{\g \mu' x_m^2 }{\e}\triangleq S_{1}(\e), }
\eqs{c21}{\frac{1}{(\sum_k \hat{\a}_k)^2}\sum_{i=1}^{N}\tau^2_i\hat{\a}_i^2\norm{\X_i\oi_i}^2_F \leq\frac{ S_{1}(\e)}{N \hat{\a}^2}.}
Since $S_{1}(\e)/\hat{\a}^2$ is $ O(\g^{2/3})$, then $\lim_{N\to\infty}  {S_1(\e)}/{N \hat{\a}^2}=0$.

Now consider the second term of $C_2$, we observe that  
\al{&\X_i\trs\oi_i\L_i\circ \X_j\trs\oi_j \L_j\\=&[\o_{i,1}\o_{j,1}c_{i,j,1}\x_{i,1}\circ \x_{j,1},\dots, \o_{i,m}\o_{j,m}c_{i,j,m}\x_{i,m}\circ \x_{j,m}],} where $$c_{i,j,k}=\frac{1}{(\s_i^2+\o_{i,k}^2)(\s_j^2+\o_{j,k}^2)}.$$
Note that $\mathbf{1}\trs(\x_{i,k}\circ \x_{i,j})=\x_{i,k}\trs\x_{i,j}$, we set $S_2=\max_{k,j}\max_{i}\x_{i,k}\trs\x_{i,j}$. 
By substituting the terms with their extreme values and the  Cauchy–Schwarz inequality, we can obtain
\al{\mathbf{1}\trs (\X_i\trs\oi_i\L_i\circ & \X_j\trs\oi_j \Lambda_j )\mathbf{1} \\
	=&S_2\sum_{k=1}^{m}\frac{\o_{i,k}\o_{j,k}}{(\s_i^2+\o_{i,k}^2)(\s_j^2+\o_{j,k}^2)}\\
	<& S_2\sum_{k=1}^{m}\frac{\o_{i,k}\o_{j,k}}{\s_i^2\s_j^2}
	<   \frac{S_2}{\e^2} \sum_{k=1}^{m}  \o_{i,k}\o_{j,k}\\=&\frac{S_2}{\e^2}\norm{\L_i\L_j}_F\leq \frac{S_2}{\e^2}\norm{\L_i}_F\norm{\L_j}_F.}

It follows that
\eqs{c22}{&\frac{1}{(\sum_l \hat{\a}_l) ^2}\sum_{k=1}^{N}\sum_{j=1}^{N}\v_k\v_j \hat{\a}_k \hat{\a}_j \mathbf{1}\trs( \X_k\trs\oi_k\L_k\circ \X_j\trs\oi_j \L_j ) \mathbf{1}\\
	\leq&\frac{S_2\g\mu'}{\e^2 (\sum_l \hat{\a}_l) ^2}\Big(\sum_{k=1}^{N}\hat{\a}_k\sqrt{\sum_{l=1}^m\o_{k,l}^2} \Big)^2
	\leq \frac{S_2\mu'\g m \o_2}{\e^2}=O(\g^{1/3}).
}
By \rf{c21} and \rf{c22}, we say that $ C_2$ is upper bounded by $O(\g^{1/3})$.

\subsubsection{Limiting property of $C_1$}

Now consider $ C_1/N$. As $C_1$ (defined in \rf{c1}) is the weighted average of $C_{1,i}$'s (defined in \rf{cts}), we only need to discuss the limiting property of $C_{1,i}$.

Consider the first term of $C_{1,i}$,  
\eqs{c11}{\tau_i^2(1-2\a_i)\norm{\X_i\trs\oi_i}^2_F&<
	\tau_i^2\norm{\X_i\oi_i}_F^2\\<&m  \mu'\g\s_i^2x_m^2\sum_{k=1}^{m}\frac{1}{(\s_i^2+\o_{i,k}^2)^2 } \\
	<&\frac{x_m^2m^2\mu'\g  }{\e} \triangleq S_3(\e).
}
The first line is obtained by eliminate the positive term $2\a_i$ from the $1-s\a_i$, and the second follows by eliminating $\o_{i,k}^2$ terms from the denominators. Note that $ S_3(\e)=O(\g^{2/3})$, and it follows that
$\lim\limits_{N\to \infty}{ S_3(\e)}/{N}=0. $

The analysis of second term of $C_{1,i}$ is the same  as in\rf{c21}.  Note that
when divided by $N$, it follows that
\[\lim\limits_{N\to \infty}\frac{ S_{1}(\e)}{N^2 \hat{\a}^2}=0,  \]

For the third part of $C_{1,i}$, we observe that
\al{\v_i^2\norm{\X_i\oi_i\L_i}_F^2=&mx_m^2 \mu'\g\sum_{k=1}^{m} \frac{\o_{i,k}^2}{(\s_i^2+\o_{i,k}^2)^2}  \\<& \frac{m^2x_m^2 \mu'\g}{\e^2}\triangleq S_4(\e).     }
Since $S_4(\e)=O(\g^{1/3})$, we have
$\lim\limits_{N\to \infty} { S_4(\e)}/{N}=0.    $

The fourth term of $C_1$ is the same as the second part of $C_2$, 
then
\[ \lim\limits_{N\to\infty} \frac{S_2\mu'\g m \o_2}{N\e^2} =0.  \]

Finally, we consider the last part of $C_1$. We set $S_5=\min_{k,j}\min_{i}\x_{i,k}\trs\x_{i,j}$ and find $M$ such that $\sqrt{M}/\e=\max_i(\s_i^2+\o_i^2)$.
\al{\mathbf{1}\trs (\X_i\trs\oi_i\L_i\circ &\X_j\trs\oi_j \Lambda_j )\mathbf{1} \\
	>&S_5\sum_{k=1}^{m}\frac{\o_{i,k}\o_{j,k}}{(\s_i^2+\o_{i,k}^2)(\s_j^2+\o_{j,k}^2)}\\
	>& S_5\sum_{k=1}^{m}\frac{\o_{i,k}\o_{j,k}}{[\max_j(\s_j^2+\o_{j,k}^2)]^2}\\
	>&\frac{S_5M}{\e^2} \sum_{k=1}^{m} \o_{i,k}\o_{j,k}=\frac{S_5M}{\e^2}\norm{\L_i\L_j}_F.}
Then 
\al{\sumbkss{2} \v_i\v_k \mathbf{1}\trs( \X_i\trs\oi_i\L_i \circ& \X_k\trs\oi_k \L_k) \mathbf{1}\\>&\sumbkss{2  S_5M/\e^2} \v_i\v_k \norm{\L_i\L_k}_F\\\geq& \frac{2 S_5\mu\g M\o_1\sqrt{m}}{\e^2 }
	\triangleq S_6(\e).
}
 Then 
$\lim\limits_{N\to \infty} { S_6(\e)}/{N}=0.    $

Since all terms of $C_1$ go to zero when divided by $N$, we conclude that 
\eq{cc}{\lim\limits_{N\to\infty}\frac{ C_1}{N}=0} 

\subsubsection{Proof of consistency}
By  \rf{c21} , \rf{c22}, and \rf{cc}, and when we choose $\l \hl \sim N$ and $\g\sim \e^4$,  then 
\al{\lim_{t\rightarrow \infty}&\lim_{N\rightarrow \infty}\E[\norm{\hz-\tw}^2]=\lim_{t\rightarrow \infty}\lim_{N\rightarrow \infty}2\E[U_t]\\ \leq&  \frac{1}{\mu\k}\Big(\frac{\eta\mu'-\k\mu}{\l\beta\hl  }C_1+C_2\Big)
< \frac{S_2\mu'\g m \o_2}{\mu\k\e^2}=O(\g^{1/3}),  }
where $S_2=\max_{k,j}\max_{i}\x_{i,k}\trs\x_{i,j}$.
\subsection{Proof of Theorem 3}\label{T3}

In this section, we first derive the formula of $d\w_t$ in the similar way as in \rf{A1}, and then provide the convergence property of the federated learning. 

\subsubsection{Continuous time representation of FL}
We rewrite the scheme \rf{fl1} as:
\eqs{022}{\w_{t}=&\w_{0}-
	\g\sum_{i=1}^N\sum_{k=1}^{N_{g,i}(t/\g)} \X_{i}\trs\oi_i\X_{i}(\w_{i,l}-\tw)\\&
	+ \g\sum_{i=1}^N\sum_{k=1}^{N_{g,i}(t/\g)} \X_{i}\trs\oi_i\ve_{i,l} +\\& \g\sum_{i=1}^N\sum_{k=1}^{N_{g,i}(t/\g)} \X_{i}\trs \oi_i\Lambda_i\xi_l  
	+\g\sum_{i=1}^N\sum_{k=1}^{N_{g,i}(t/\g)} \e_{i,l}.  }

By \rf{03},   and \rf{06},  we can obtain
\eqs{fl2}{d\w_t=&\sum_{i=1}^{N}\Big[-\mu_i g\itt  dt +\tau_i\X_i\trs \oi_i  dB\itt \\&+\v_i\X_{i}\trs\O_i^{-1}\L_i d B_t\Big].  }

\subsubsection{Proof of Theorem 3}
The proof follows a similar outline as Theorem 1. 
We apply Ito's lemma to $dF_t$ to get the following form,
\al{dF_t=&(\w_t-\tw)\cdot d(\w_t-\tw)+\\&\frac{1}{2}d(\w_{t}-\tw)\cdot d(\w_t-\tw)\\
	=&-\sum_{i=1}^{N}\mu_i g\itt \trs(\w_t-\tw)dt+\\& \sum_{i=1}^{N}\tau_i\big( \X_i\trs\oi_idB\itt   \big)\trs(\w_t-\tw)       \\&+ \sum_{i=1}^{N}\v_i(\X_i\trs\oi_i\L_i dB_t)\trs(\w_t-\tw)  +   \\& \frac{1}{2}\Big(\sum_{i=1}^{N}\tau_i^2\norm{ \X_i\trs\oi_i}_F^2+
	\sum_{k=1}^N\sum_{j=1}^{N}\v_k\v_j A_{k,j}  \Big)dt.}
We rewrite it as follows, 
\al{dF_t=&  -\sum_{i=1}^{N}\mu_ig\itt\trs(\w_t-\tw) dt+K_3d\td{B}_{f,t}+C_3 dt,   }
where $K_3d\td{B}_{f,t}$ is the summation of the Ito terms, and $C_3$ is the summation of the constant terms: 
\al{K_3d\td{B}_{f,t}=& \sum_{i=1}^{N}\tau_i(\w_t-\tw)\trs\X_i\trs \oi_i   dB\itt \\&+ \sum_{i=1}^{N}\v_i(\w_t-\tw)\trs\X_{i}\trs\O_i^{-1}\L_i d B_t,} and
\al{ C_3=&\frac{1}{2} \big(  \sum_{i=1}^{N}\tau_i^2\norm{ \X_i\trs \oi_i }_F^2+\sum_{k=1}^N\sum_{j=1}^{N}\v_k\v_jA_{k,j}   \big).     }
Here, we note that the quadratic variation of $\w_t-\w^*$ is defined as, 
\[ \vb{\w_t-\w^*}_{t}=\E\int_0^{t} C_3 ds\to \var(\w_t-\w^*).\]
Hence $tC_3$ describes the variation of $\w_t-\w^*$.
Note that 
	$$(\tr f(x_1)-\tr f(x_2))^T(x_1-x_2)\geq \frac{\k}{2}\norm{x_1-x_2}^2,$$ then
\al{-\sum_{i=1}^{N}\mu_ig\itt\trs&(\w_t-\tw) dt\\
	\leq& -\mu\sum_{i=1}^{N}(g\itt-g(\w^*))\trs(\w_t-\tw) dt\\
	\leq &-\k\mu \norm{\w_t-\tw}^2 dt.
    }
    It follows that
\[d F_t\leq -2\mu\k F_t dt +K_3d\td{B}_{f,t}+C_3 d t.\]
Now consider $d(e^{2\k \mu t}F_t)$,
\al{ d(e^{2\k \mu t}F_t) =& e^{2\k \mu t}dF_t+ 2\k \mu e^{2\k \mu t}F_t  dt  \\
	\leq& e^{2\k \mu t}C_3dt + e^{2\k \mu t}K_3d\td{B}_t.
}
Integrating the left and right hand side of the inequality above yields  
\al{F_t\leq&  e^{-2\k \mu t}\Big[F_0    + \int_{0}^{t}e^{2\k \mu s}C_3ds +  \int_{0}^{t} e^{2\k \mu s}K_3d\td{B}_s\Big]\\
	= &e^{-2\k \mu t}F_0    +\frac{1}{2\k\mu} C_3\Big(1-e^{-2\k \mu t}\Big) \\&  +  e^{-2\k \mu t}\int_{0}^{t} e^{2\k \mu s}K_3d\td{B}_s.}
Finally, when we take the expectation of the above inequality, by observing that the last term is a martingale we get 
\al{\E[F_t]\leq e^{-2\k \mu  t}F_0+\frac{1}{2\k\mu}  C_3(1-e^{-2\k \mu t}).}
Now, we consider the lower bound of $F_t$. By Lipschitz continuity, 
\al{-\sum_{i=1}^{N}\mu_ig\itt\trs&(\w_t-\tw) dt\\
	=& -\sum_{i=1}^{N}\mu_i(g\itt-g(\w^*))\trs(\w_t-\tw) dt\\
	\geq& -\mu'\norm{\sum_{i=1}^{N}g\itt-g(\w^*)}\norm{\w_t-\tw} dt\\
	\geq& -\eta\mu'\norm{\w_t-\tw}^2 dt}
    It follows that
\[d F_t\geq -2\eta\mu'F_t dt +K_3d\td{B}_{f,t}+C_3 d t.\]
Now consider $d(e^{2\eta \mu' t}F_t)$,
\al{ d(e^{2\eta \mu' t}F_t) =& e^{2\eta \mu' t}dF_t+ 2\eta \mu' e^{2\eta \mu' t}F_t  dt  \\
	\geq& e^{2\eta \mu' t}C_3dt + e^{2\eta\mu'  t}K_3d\td{B}_t.
}
It gives that
\al{F_t\geq&  e^{-2\eta \mu' t}F_0    + \int_{0}^{t}e^{2\eta \mu' s}C_3ds +  \int_{0}^{t} e^{2\eta \mu' s}K_3d\td{B}_s\\
	= &e^{-2\eta \mu' t}F_0    +\frac{1}{2\eta \mu'} C_3\Big(1-e^{-2\eta \mu' t}\Big)   + \int_{0}^{t} e^{2\eta \mu' s}K_3d\td{B}_s.}
Finally, taking expectation of both sides yields 
\al{\E[F_t]\geq  &e^{-2\eta \mu' t}F_0+\frac{1}{2\eta \mu'}  C_3(1-e^{-2\eta \mu' t}).  }



\bibliography{mybib}

\bibliographystyle{ieeetr}

\end{document}